\documentclass[10pt,journal,cspaper,compsoc]{styles/IEEEtran}

\ifCLASSOPTIONcompsoc
\else
\fi

\ifCLASSINFOpdf
  \usepackage[pdftex]{graphicx}
  \graphicspath{{figures/}}
  \DeclareGraphicsExtensions{.pdf,.jpeg,.png}
\else
\fi

\usepackage{color,soul}

\ifCLASSOPTIONcompsoc
\usepackage[tight,normalsize,sf,SF]{subfigure}
\else
\usepackage[tight,footnotesize]{subfigure}
\fi

\usepackage{color}
\usepackage{multirow}

\newcommand{\vottoolkitpage}{https://github.com/vicoslab/vot-toolkit}
\newcommand{\first}[1]{%
    \bf{\color{red}#1}%
}
\newcommand{\second}[1]{%
    \em{\color{blue}#1}%
}
\newcommand{\third}[1]{%
    {\color{green}#1}%
}

\usepackage[colorinlistoftodos]{todonotes}
 
\begin{document}
 
\title{A Novel Performance Evaluation Methodology for Single-Target Trackers}
 
\author{Matej~Kristan,~\IEEEmembership{Member,~IEEE,}
         Jiri~Matas, Ale{\v s}~Leonardis,~\IEEEmembership{Member,~IEEE,} Tom{\' a}{\v s}~Voj{\'i}{\v r}, Roman~Pflugfelder, Gustavo~Fern{\' a}ndez, Georg~Nebehay, Fatih~Porikli and Luka~{\v Cehovin}~\IEEEmembership{Member,~IEEE,}
\IEEEcompsocitemizethanks{\IEEEcompsocthanksitem M. Kristan and L. {\v Cehovin} are with the Faculty of Computer and Information Science, University of Ljubljana, Slovenia.\protect\\
E-mail: see http://www.vicos.si/People/Matejk
\IEEEcompsocthanksitem R. Pflugfelder, G. Nebehay and G. Fern{\' a}ndez are with  Austrian Institute of Technology, Austria
\IEEEcompsocthanksitem J. Matas and T. Voj{\'i}{\v r} are with the Faculty of Electrical Engineering, Czech Technical University in Prague, Czech Republic
\IEEEcompsocthanksitem A. Leonardis is with the University of Birmingham, School of Computer Science, United Kingdom
\IEEEcompsocthanksitem F. Porikli is with NICTA and Australian National University, Australia}
\thanks{}}

\markboth{IEEE TRANSACTIONS ON PATTERN ANALYSIS AND MACHINE INTELLIGENCE, (accepted for publication) 2016}%
{Kristan \MakeLowercase{\textit{et al.}}}

\IEEEcompsoctitleabstractindextext{
\begin{abstract}
This paper addresses the problem of single-target tracker performance evaluation. We consider the performance measures, the dataset and the evaluation system to be the most important components of tracker evaluation and propose  requirements for each of them. The requirements are the basis of a new evaluation methodology that aims at a simple and easily interpretable tracker comparison. The ranking-based methodology addresses tracker equivalence in terms of statistical significance and practical differences. A fully-annotated dataset with per-frame annotations with several visual attributes is introduced. The diversity of its visual properties is maximized in a novel way by clustering a large number of videos according to their visual attributes. This makes it the most sophistically constructed and annotated dataset to date. A multi-platform evaluation system allowing easy integration of third-party trackers is presented as well. The proposed evaluation methodology was tested on the VOT2014 challenge on the new dataset and 38 trackers, making it the largest benchmark to date. Most of the tested trackers are indeed state-of-the-art since they outperform the standard baselines, resulting in a highly-challenging benchmark. An exhaustive analysis of the dataset from the perspective of tracking difficulty is carried out. To facilitate tracker comparison a new performance visualization technique is proposed.
\end{abstract}
 
}

\maketitle

\IEEEdisplaynotcompsoctitleabstractindextext
 
\IEEEpeerreviewmaketitle

\section{Introduction}

Visual tracking is a rapidly evolving field that has been increasingly attracting attention of the vision community.  It offers many scientific challenges and it emerges in other computer vision problems such as motion analysis, event detection and activity recognition. A steady increase of hardware performance and its price reduction have opened a vast application potential for tracking algorithms including surveillance systems, automotive systems, transport, sports analytics, medical imaging, mobile robotics, film post-production and human-computer interfaces.

The activity in the field is reflected in abundance of new tracking algorithms presented in journals and at conferences summarized in the many survey papers, e.g.,~\cite{Gavrila99,Moeslund2001,Gabriel03,Hu2004,Moeslund2006,Yilmaz2006,Li2013}. However, the boom in tracker proposals has not been accompanied by standardization of the methodology for their objective comparison.

One of the most influential performance analysis efforts for object tracking is PETS (Performance Evaluation of Tracking and Surveillance)~\cite{Young2005}. The first PETS workshop took place in 2000 aiming at evaluation of visual tracking algorithms for surveillance applications. Its focus gradually shifted to high-level event interpretation algorithms. Other frameworks and datasets have been presented since, but these focused on evaluation of surveillance systems and event detection, e.g., CAVIAR\footnote{http://homepages.inf.ed.ac.uk/rbf/CAVIARDATA1}, i-LIDS~\footnote{http://www.homeoffice.gov.uk/science-research/hosdb/i-lids}, ETISEO\footnote{http://www-sop.inria.fr/orion/ETISEO}, change detection~\cite{GoyetteCVPR12}, sports analytics (e.g., CVBASE\footnote{http://vision.fe.uni-lj.si/cvbase06/}), specialized on tracking specific objects like faces, e.g., FERET~\cite{Phillips2000}, ~\cite{Kasturi_TPAMI_2009} or tracking for autonomous vehicles, e.g., KITTI~\cite{Geiger2012CVPR}. Recently, several works have been published in the broad area of model-free visual object tracking evaluation, eg.,~\cite{Bernardin2008,Karasulu2011,Salti2012,Wu2013,Smeulders2013,Pang2013,Kristan2013,Kristan2014} and following the success of the VOT challenges~\cite{Kristan2013,Kristan2014} a performance evaluation benchmark for multiple target tracking was presented as well~\cite{MOTChallenge2015} .

There are several important subfields in visual tracking, ranging from multi-camera, multi-target~\cite{Bernardin2008,Fleuret08a,MOTChallenge2015}, to single-target~\cite{Wu2013,Smeulders2013,Pang2013,Kristan2013,Kristan2014} trackers. These subfields are quite diverse, without a unified evaluation methodology and specific methodologies have to be tailored to each subfield.

In this paper, single-camera, single-target, model-free, causal trackers, applied to short-term tracking are considered. The model-free property means that the only supervised training example is provided by the bounding box in the first frame. The short-term tracking means that the tracker does not perform re-detection after the target is lost. Drifting off the target is considered a failure. The causality means that the tracker does not use any future frames to infer the object position in the current frame. The tracker output is specified by a rotated bounding box.

\subsection{Requirements for tracker evaluation}\label{sec:Requirements}

The evaluation of new tracking algorithms depends on three essential components: (1)~performance evaluation measures, (2)~a dataset and (3)~an evaluation system. In the following, the requirements for these components are stated.

\textbf{Performance measures.} A wealth of performance measures have been proposed for single-object tracker evaluation, but there is no consensus on which measure should be preferred. Ideally, measures should clearly reflect different aspects of tracking. Apart from merely ranking, we also need to determine cases when two or more trackers are performing equally well. We require the following: The measures should allow an easy interpretation and should support tracker comparison with a well-defined {\em tracker equivalence}.

\textbf{Datasets.} The dataset should allow evaluation of trackers under diverse conditions like partial occlusion, clutter and illumination changes. One approach is to construct a very large dataset, but this does not guarantee diversity in visual attributes and it significantly slows down the process of evaluation. A better approach is to annotate each sequence with the visual attributes occurring in that sequence and perform clustering to reduce the size of the dataset, while keeping it diverse. Annotation is also important for per-attribute tracker analysis. A common approach is to annotate a sequence globally with an attribute if that attribute occurs anywhere in the sequence. The trackers can then be compared only on the sequences corresponding to a particular attribute. However, visual phenomena do not usually last throughout the entire sequence. For example, a partial occlusion might occur at the end of a sequence, while a tracker might fail due to some other effects occurring at the beginning of the sequence. In this case, the failure would be falsely attributed to the occlusion. A per-frame dataset labeling is thus required to facilitate a more precise analysis. This motivates the following requirements: (1) The dataset should be diverse in visual attributes. (2) Per-frame annotation of visual attributes is required.

\textbf{Evaluation systems.} For a rigorous evaluation, an evaluation system that performs the same experiment on different trackers using the same dataset is required. A wide-spread practice is to initialize the tracker in the first frame and let it run until the end of a sequence. However, the tracker might fail right at the beginning of the sequence due to some visual degradation, effectively meaning that the system utilized only the first few frames for evaluation of this tracker. Thus the first requirement for the system is that it fully uses the data. This means that once the tracker fails, the system has to detect the failure and reinitialize the tracker. Therefore, a certain level of interaction, that goes beyond simple running until the end of the sequence, is required. Furthermore, the evaluation system has to also account for the fact that the trackers are typically coded in various programming languages and often platform-dependent. This motivates the following set of requirements the evaluation system should meet: (1) Full use of the dataset. (2) Allow interaction with the tracker. (3) Support for multiple platforms. (4) Easy integration with trackers.

\subsection{Our contributions}\label{sec:Contributions}

In this paper we present the following four contributions:

The first contribution is a novel tracker evaluation methodology based on two simple, easy interpretable, performance measures. A significant novelty of the proposed methodology is the use and first of its kind analysis of reinitializations at tracking failures. Reinitialization-based measures are compared theoretically and experimentally to standard counterparts that do not apply reinitialization. We propose a first of its kind tracker ranking methodology that addresses the concept of tracker equivalence and takes into account statistical significance as well as practical difference in tracking accuracy. A new visualization of ranks is proposed as well to aid comparative analysis. 

The second contribution is a new dataset and evaluation system. The dataset is constructed by a novel video clustering approach based on visual properties. The dataset is fully annotated, all the sequences are labeled per-frame with visual attributes to facilitate in-depth analysis. The benefits of per-frame attribute annotation are analyzed theoretically and experimentally. The proposed evaluation system enjoys multi-platform compatibility and offers easy integration with trackers. The system has been tested in a large-scale distributed experiment on the VOT2013 and VOT2014 challenges.

The third contribution is a detailed comparative analysis of 38 trackers using the proposed methodology, making it the largest benchmark to date.

The forth contribution is a novel analysis of the sequences in the dataset from the perspective of tracking success.

Preliminary versions of some parts of this paper have been previously published (during the period 2013-2014) in three workshop papers~\cite{Kristan2013a,Kristan2014,KRISTAN_2014_ECCV}.

The remainder of the paper is structured as follows: In Section~\ref{sec:relatedWork} the most related work is reviewed and discussed. The new tracker evaluation methodology is presented and theoretically analyzed in Section~\ref{sec:evalMethodology}, while the new dataset selection approach, the evaluation system and the results of the experimental analysis are presented in Section~\ref{sec:experiments}. Conclusions are drawn in Section~\ref{sec:Conclusion}.

\section{Related work}\label{sec:relatedWork}

\subsection{Performance measures}

A wealth of performance measures have been proposed for single-object tracker evaluation. These range from basic measures like center error~\cite{Ross2008}, region overlap~\cite{Li2011}, tracking length~\cite{Kwon2009}, failure rate~\cite{Kristan2008b,Kristan2010b}, F-score~\cite{Karasulu2011,Smeulders2013}, pixel-based precision~\cite{Karasulu2011}, to more sophisticated measures, such as CoTPS~\cite{Nawaz2013} \cite{Carvalho2012}, which combine several measures. A nice property of the combined measures is that they provide a single score to rank the trackers. A downside is that they offer little insight into the tracker performance which  limits their interpretability. All measures strongly depend on the experimental setup within which they are computed. For example, some evaluation protocols, like Wu et al.~\cite{Wu2013} and Smeulders et al.,~\cite{Smeulders2013} initialize the trackers at the beginning of the sequence and let them run until the end. Measures computed in such a setup are inappropriate for short-term tracking evaluation, since the trackers are not expected to perform re-detection. The values of performance measures thus become irrelevant after the point of tracking failure. Including the frames past the point of failure in the computation of a global performance measure introduces significant distortions since failures closer to the beginning of the sequence are significantly  more penalized than failures occurring later in the sequence.

\begin{figure} 
    \centering
    \includegraphics[width=\columnwidth]{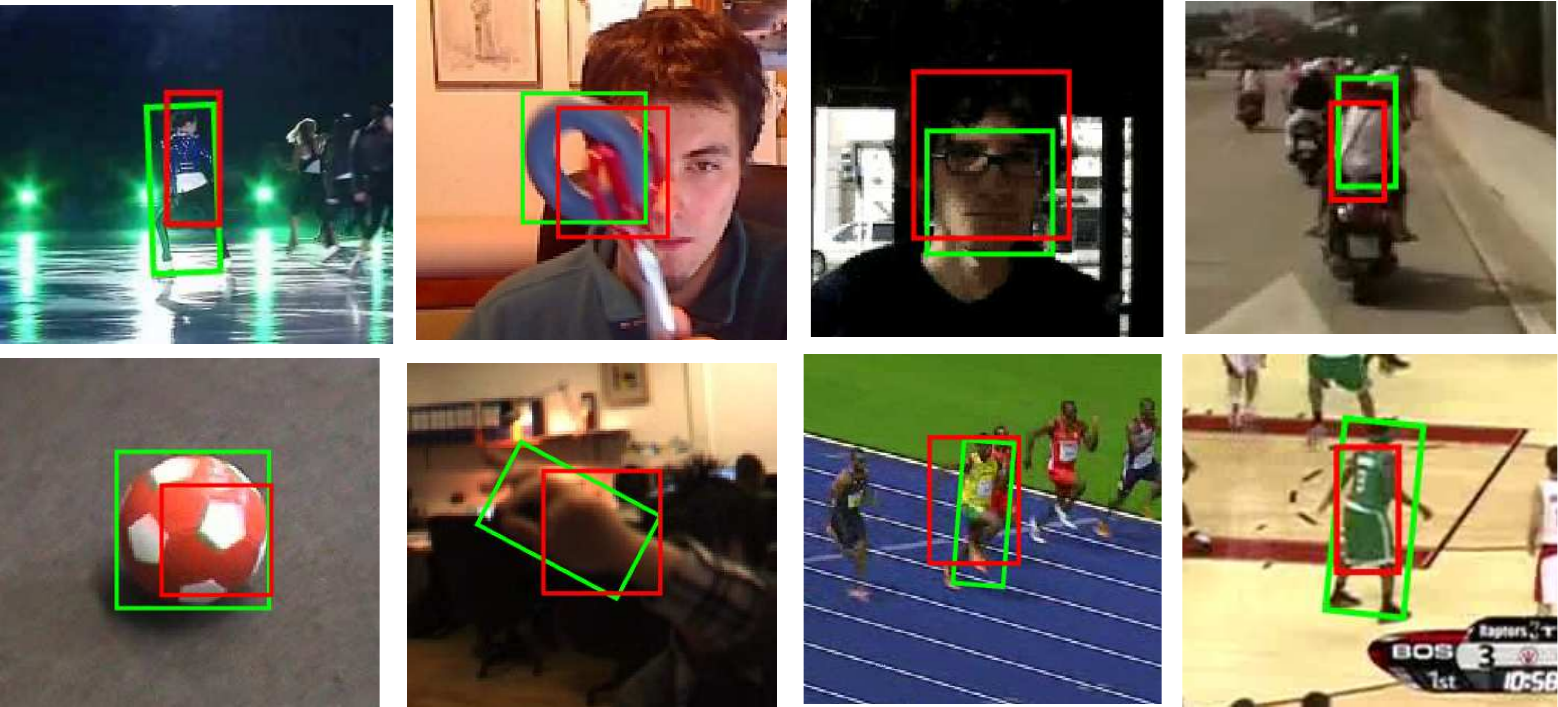}
    \caption{Examples of bounding boxes (red) at 0.5 overlap with the ground truth (green). Notice that the rectangles still fit the objects quite well.}
    \label{fig:overlap_half}
\end{figure}

While some authors choose several basic measures to compare their trackers, recent studies~\cite{Cehovin2014wacv,CehovinTracMeas2015} have shown that many measures are correlated and do not reflect diverse aspects of tracking performance. In this respect, choosing a large number of measures may in fact again bias results toward some particular aspects of tracking performance. Smeulders et al.~\cite{Smeulders2013} propose using two measures: an F-score calculated at the Pascal region overlap criterion~(threshold~$0.5$)~\cite{Everingham2014} and a center error. Note that the F-score based measure was originally designed for object detection. The threshold~$0.5$ is also rather high and there is no clear justification of why exactly this threshold should be used to compare trackers~\cite{Wu2013} since it is hardly an indicator of tracking failure (see examples in Figure~\ref{fig:overlap_half}).

Since the center error becomes arbitrary high once the tracker fails, Wu et al.~\cite{Wu2013} propose to measure the percentage of frames in which the center distance is within some prescribed threshold. However, this threshold significantly depends on the object size, which makes this particular measure quite brittle. A normalized center error measured during successful tracks may be used to alleviate the object size problem, however, the results in~\cite{Smeulders2013} show that the trackers do not differ significantly under this measure which makes it less appropriate for tracker comparison. As an additional measure,~\cite{Wu2013} propose an area under a ROC-like plot of thresholded overlaps. Recently,~\cite{Cehovin2014wacv} have shown that this is equivalent to the average region overlap measure computed from all frames of sequences. In fact, based on an extensive analysis of performance measures, {\v C}ehovin et al.
~\cite{CehovinTracMeas2015}
argue that the region overlap is superior to the center error.

While it is important to study and evaluate the tracker performance separately in terms of several less correlated performance measures, it is sometimes required to rank trackers in a single rank list. In this case a convenient strategy is to combine these measures into {\em rank averaging}, similarly to what was done in the change detection challenge~\cite{GoyetteCVPR12}. In rank averaging, competing algorithms are ranked with respect to several performance measures and their ranks are averaged. This simulates competition of trackers with respect to different performance measures and assumes equal importance of all measures. The fact that trackers are ranked along each measure induces normalization of measures to a common scale prior to averaging.

\subsubsection{Visual performance evaluation}

Several authors propose to visually compare tracking performance via performance summarization plots. These plots show the percentage of frames for which
the estimated object location is within some threshold distance of the ground truth. Most notable are precision plots \cite{Yilmaz2006,babenko2011tpami,Wu2013}, which measure the object location accuracy in terms of center error. Alternatively, success plots~\cite{Salti2012,Wu2013} use the region overlap instead. Salti et al.,~\cite{Salti2012} implicitly account for variable threshold dependency by plotting the percentage of correctly tracked frames with respect to the mean region overlap within these frames. {\v C}ehovin et al.~\cite{Cehovin2014wacv,CehovinTracMeas2015} propose a similar visualization, but they apply a single, zero, threshold on the overlap. A tracker is thus represented as a single point in this 2D space, rather than a curve, which allows easier comparison. A drawback of performance plots is that they typically become cluttered when comparing several trackers on several sequences in the same plot. To address this, Smeulders et al.~\cite{Smeulders2013} calculate a performance measure per sequence for a tracker and order these values from highest to lowest, thus obtaining a so-called survival curve. The performance of several trackers is then compared on the entire dataset by visualizing their survival curves.

\subsection{Datasets}

It is a common practice to compare trackers on many publicly-available sequences, which have became a de-facto standard in evaluation of new trackers. However, many of these sequences lack a standard ground truth labeling, which makes comparison of algorithms difficult. To sidestep this issue, Wu et al.~\cite{Wu_TPAMI2010} have proposed a protocol for stochastic tracker evaluation on a selected dataset that does not require ground truth labels. A similar approach was adapted by~\cite{SanMiguel2012} to evaluate tracking algorithms on long sequences. Datasets with various visual phenomena equally represented are not usually used.  In fact, many popular sequences are conceptually similar, which makes the results biased toward some particular types of the phenomena. To address this issue, Wu et al.~\cite{Wu2013} annotated each sequence with several visual attributes and report tracker performance with respect to each attribute separately. However, a per-frame annotation is not provided and not all sequences are in color, which makes results skewed with proportions of color and gray sequences. Recently, Smeulders et al.~\cite{Smeulders2013}, have presented a very large dataset called `Amsterdam Library of Ordinary Videos'~(ALOV). The dataset is composed of over three hundred sequences collected from published datasets and additional YouTube videos. The sequences are assigned to one of thirteen classes of difficulty~\cite{Chu2010} and, with the exception of ten long sequences, are kept short to increase the diversity. The sequences are not annotated per-frame with visual attributes, some sequences contain cuts and ambiguously defined targets such as fireworks which makes the dataset inappropriate for short-term tracking evaluation.

\subsection{Evaluation systems}

The most notable and general evaluation systems are ODViS~\cite{Jaynes2002}, VIVID~\cite{Collins2005}, ViPER~\cite{Doermann2000}. The former two focus on the design of surveillance systems, while the latter is a set of utilities/scripts for annotation and computation of different types of performance measures.
The recently proposed ViCamPEv~\cite{Karasulu2011} toolkit is dedicated to testing a pre-determined set of OpenCV-based basic trackers. None of these systems support interaction with the tracker, which limits their applicability. Collecting the results from the existing publications is an alternative for benchmarking trackers. Pang et al.~\cite{Pang2013} have proposed a page-rank-like approach to data-mine the published results and compile unbiased ranked performance lists. However, as the authors state in their paper, the proposed protocol is not appropriate for creating ranks of the recently published trackers due to the lack of sufficiently many publications that would compare these trackers.

\section{Visual object tracker evaluation}\label{sec:evalMethodology}

The proposed methodology assumes that the evaluation system and the dataset fulfill the requirements stated in Section~\ref{sec:Requirements}, i.e., (i)  the dataset is per-frame annotated by visual attributes and the object positions are denoted by possibly rotated bounding boxes, (ii) trackers are run on each sequence of the dataset. Once the tracker drifts off the target, the system detects a tracking failure and re-initializes the tracker. All trackers are run multiple times to account for their possibly stochastic nature.

\subsection{Evaluation methodology}\label{sec:vot_eval_met}

Based on the recent analysis of widely-used performance measures
~\cite{Cehovin2014wacv,CehovinTracMeas2015} two weakly-correlated and easily interpretable measures were chosen: (i)~accuracy and (ii)~robustness. The accuracy at time-step $t$ measures how well the bounding box~$A_t^T$ predicted by the tracker overlaps with the ground truth bounding box~$A_t^G$ and is defined as the intersection-over-union
\begin{equation}\label{eq:overlap}
 \phi_t = \frac{A_t^G \cap A_t^T}{A_t^G \cup A_t^T}.
\end{equation}
The robustness is the number of times the tracker failed, i.e., drifted from the target, and had to be reinitialized. A re-initialization is triggered when the overlap (Eq. \ref{eq:overlap}) drops to zero.

The re-initialization of trackers might introduce a bias into the performance measures. If a tracker fails at a particular frame, e.g., due to occlusion, it will likely fail again immediately after re-initialization. To reduce this bias, the tracker is re-initialized $N_\mathrm{skip}=5$ frames after the failure. The reasoning behind the choice of this value is that short-term occlusions do not last for more than five frames and we provide experimental study of this parameter in  Section~\ref{sec:skipping_exp} for completeness. In the case of a full occlusion, the tracker is initialized on the first frame in which the object is not fully occluded. A similar bias occurs in the accuracy measure. The overlaps in the frames right after the initialization are biased towards higher values over several frames and it takes a few frames of the burn-in period to reduce the bias. This means that we label the frames in the burn-in period as invalid and do not use them in computation of the accuracy. In Section~\ref{sec:burnin_exp} a study is reported in which we measured the time it takes for the overlap to approximately stabilize after reinitialization. According to the results of that study, the burn-in period is set to $N_\mathrm{burnin}=10$ frames\footnote{Note that the burn-in period would in principle depend on the frame rate as well as the speed at which an object moves. Sequences from our dataset are not recorded at high-speed and are taken at approximately the same frame rate, so the burn-in period of ten frames is a reasonable choice to remove the reinitialization bias.}.
 
A tracker is run on each sequence $N_\mathrm{rep}$ times which allows dealing with the potential variance of its performance. In particular, let $\Phi_t(i,k)$ denote the accuracy of $i$-th tracker at frame $t$ at experiment repetition $k$. The per-frame accuracy is obtained by taking the average over these, i.e., $\Phi_t(i) = {1 \over N_\mathrm{rep}}\sum\nolimits_{k=1}^{N_\mathrm{rep}}{\Phi_t(i,k)}$. The average accuracy of the $i$-th tracker, $\rho_\mathrm{A}(i)$, over some set of $N_\mathrm{valid}$ valid frames is then calculated as the average of per-frame accuracies
\begin{equation}\label{eq:avaccuracy}
\rho_\mathrm{A}(i)={1 \over N_\mathrm{valid}}\sum\nolimits_{j=1}^{N_\mathrm{valid}}{\Phi_j(i)}.
\end{equation}

In contrast to accuracy measurements, a single measure of robustness per experiment repetition is obtained. Let~$F(i,k)$ be the number of times the $i$-th~tracker failed in the experiment repetition~$k$ over a set of frames. The average robustness of the $i$-th~tracker is then
\begin{equation}\label{eq:avrobustness}
\rho_R(i) = {1 \over N_\mathrm{rep}}\sum\limits_{k=1}^{N_\mathrm{rep}}F(i,k).
\end{equation}

The overall performance on the dataset can be estimated as the weighted average of the per-sequence performance measures, with weights proportional to the lengths of the sequences. Note that this is equivalent to concatenating the sequence of per-frame overlaps/failures from the entire dataset into a single super-sequence and calculating the two averages in (\ref{eq:avaccuracy}) and (\ref{eq:avrobustness}). Similarly, per-visual-attribute performance can be evaluated for a specific attribute by collecting all the frames labelled as that attribute into an attribute super-sequence and calculating (\ref{eq:avaccuracy}) and (\ref{eq:avrobustness}).

For a fair comparison, we propose a ranking-based methodology akin to~\cite{Everingham10,GoyetteCVPR12} but we introduce the concept of equally-ranked trackers. For each tracker, a group of so-called equivalent trackers containing trackers performing indistinguishably is determined and a corrected rank is then calculated. There are several choices for calculating the correction, e.g., one could take the min, max or mean of ranks in the group. The least conservative choice is max, since it always penalizes a tracker if the equivalency test cannot confirm the difference from a lower-ranked tracker, and on the other hand, the min is most conservative, since it always makes a correction in interest of the tracker. In the subsequent evaluation we use the mean of the ranks as a compromise between the two extrema. Note that the concept of equivalent trackers is not transitive, and should not be mistaken for the standard equivalence relation. For example, consider trackers~$T_1$,~$T_2$ and~$T_3$.  It may happen that a tracker~$T_2$ performs indistinguishably from~$T_1$ and~$T_3$, but this does not necessarily mean that~$T_1$ performs equally well as both,~$T_2$ and~$T_3$. The equality of trackers should therefore be established for each tracker separately. Two types of tests for establishing performance equivalence are considered in the following.

\subsubsection{Tests of statistical differences}

A per-frame accuracy is available for each tracker. One way to gauge equivalence in this case is to apply a paired test to determine whether the difference in accuracies is statistically significant. When the differences are distributed normally, the Student's t-test, which is often used in the aeronautic tracking research~\cite{Yaakov01}, is the appropriate choice. However, in a preliminary study we have applied Anderson-Darling tests of normality~\cite{AndersonDarling1952} and have observed that the accuracies in frames are not always distributed normally, which might render the t-test inappropriate. As an alternative, the Wilcoxon Signed-Rank test as in~\cite{Demsar2006} is applied that tests a null hypothesis that differences come from a distribution with zero median (see~\cite{Navidi2011} for further details).
 
In case of robustness, several measurements of the number of tracker failures over the entire sequence in different runs is obtained. However, these cannot be paired, and the Wilcoxon Rank-Sum (also known as Mann-Whitney U-test)~\cite{Demsar2006} is used instead to test the difference in the average number of failures. This is a two-sided rank sum test which tests the null hypothesis that the number of failures of two trackers are independent samples from distributions with equal medians (see~\cite{Navidi2011} for further details).

\subsubsection{Tests of practical differences}\label{sec:prac_difference}

Note that statistical difference does not necessarily imply a practical difference~\cite{Demsar2008}, which is particularly important in equivalency tests for accuracy. The practical difference is a level of difference in accuracy that is considered negligibly small. This level can come from the noise in annotation, the fact that multiple ground truth annotations of bounding boxes might be equally valid, or simply from the fact that very small differences in tracking accuracy are negligible from a practical point of view. Therefore, a pair of trackers is considered to perform equally well in accuracy if their difference in performance is not statistically significant or if it fails the practical difference test.

In terms of practical difference, a pair of trackers $i$ and $j$ is said to perform differently if the difference of their averages is greater than a predefined threshold $\gamma$, i.e., $|\rho_\mathrm{A}(i)-\rho_\mathrm{A}(j)| > \gamma$, or, by defining a difference at frame $t$, $d_t(i,j)= \phi_t(i) - \phi_t(j)$, expanding the sums and pulling the threshold into the summation, ${1 \over T} |\sum\nolimits_{t=1}^T d_t(i,j)/\gamma| > 1$. Since the frames in the super-sequence come from multiple sequences, the thresholds $\gamma$  may vary over the frames. A pair of trackers therefore passes the test for the practical difference in accuracy if the following relation holds
\begin{equation}\label{eq:practDiff}
{1 \over T} |\sum\nolimits_{t=1}^T d_t(i,j)/\gamma_t| > 1,
\end{equation}
where $\gamma_t$ is the practical difference threshold corresponding to the $t$-th frame.

\subsubsection{Visualization of results}

Results can be visualized by the accuracy-robustness plots proposed by~\cite{Cehovin2014wacv} in which a tracker is presented as a point in terms of accuracy and robustness. The accuracy is defined as in (\ref{eq:avaccuracy}), while the robustness is converted into a probability of tracker failing after $S$ frames, thus scaling robustness into the range between zero and one. Since we have extended the methodology of~\cite{Cehovin2014wacv} to rankings, we also extend the visualization. In particular, the rank results can be displayed using the accuracy-robustness (AR) rank plots. Since each tracker is presented in terms of its rank with respect to robustness and accuracy, we can plot it as a single point on the corresponding 2D AR-rank plot.  Trackers that perform well relative to the others are positioned in the top-right part of the plot, while the, relatively speaking, poorly-performing trackers occupy the bottom-left part.

\subsection{Theoretical comparison to related works}\label{sec:theoreticalEval}

The most related works to the performance evaluation methodology presented in this paper are the methodologies presented by Wu et al.~\cite{Wu2013} and Smeulders et
al.~\cite{Smeulders2013}. In principle, all the methodologies use global averages based on the overlaps of tracker bounding boxes and ground truth.
The main difference between~\cite{Wu2013} and~\cite{Smeulders2013} is that~\cite{Wu2013} computes the average-overlap-based measure (like our approach), while~\cite{Smeulders2013} computes an F-score at 0.5 overlap. For short-term tracking, the tracker is not required to re-detect the target after losing it. This means that the tracker is not required to report the target loss and the F-score from~\cite{Smeulders2013} reduces to precision, i.e., the ratio of frames in which the overlap with ground truth is grater than 0.5. Applying such a high threshold reduces the strength of the performance measure. For example, consider a pair of trackers, tracker A and B: tracker A performs at 0.47 overlap, whereas tracker B performs at 0.1 overlap and none of the trackers ever drifts off the target. The F-score at overlap 0.5 is zero for both trackers, meaning that the measure cannot discern the performance among the trackers since their overlap is below 0.5. Furthermore, the measure would induce a large distinction between trackers A (F-score 0) and a tracker that performs at overlap 0.5 (F-score 1) even though the difference between both is only 0.03 overlap.

There are three notable differences between our methodology and~\cite{Wu2013,Smeulders2013}. The first difference is that our methodology detects tracking failure and applies re-initializations, while the~\cite{Wu2013} and~\cite{Smeulders2013} do not re-initialize, nor detect a failure. The methodology from~\cite{Wu2013} relies on compensating for this drawback by increasing the number
of sequences to 50 and recently~\cite{Smeulders2013} proposed using over 300 sequences. The second difference is that our methodology is based on per-frame visual-attribute annotation for per-visual
attribute performance evaluation. On the other hand,~\cite{Wu2013,Smeulders2013} globally annotate a sequence with all the appearing tributes. Per-visual attribute performance is then computed by using all frames of the sequences globally annotated by a particular attribute. The last difference relates to the ability to state that one tracker performs better than another. While all three methodologies produce ranks, only our methodology accounts for the practical as well as statistical difference and takes into account the noise in ground truth annotation to gauge equivalence of trackers.

The aim of the methodologies is to estimate the tracker overall or per-visual attribute performance and rank trackers according to this estimate. In this respect, the methodologies can be thought of as state estimators in which the hidden state is the tracker true performance (e.g., expected overlap). Thus, methodologies can be studied from the perspective of bias and variance of state estimators. In the following we apply this view to further analyze the properties of estimators in terms of applying re-initialization as well as per-frame visual attribute annotation.

\subsubsection{The importance of re-initialization}\label{sec:TheoryReinit}

To establish some theoretical results on performance evaluation with or without applying re-initializations, the following thought experiment is considered. Assume a tracker is tested on a set of $N$ sequences, each $N_\mathrm{s}$ frames long. A sequence $j$ contains a critical point at the frame $\alpha_j N_\mathrm{s}$, where a tracker fails with probability $p$, i.e., it drifts and remains off the target for the remaining part of the sequence. During a successful period of tracking, the per-frame overlaps are sampled from a distribution with mean $\mu_A$ and variance $\sigma_A^2$. After the failure, the overlaps fall to zero, i.e., they are sampled from a distribution with $\mu_b=0$ and $\sigma_B^2=0$. A critical point can occur anywhere in the sequence with equal probability, meaning that these points are distributed uniformly along the sequence, i.e., $\alpha_j \sim \mathcal{U}(0,1)$. A tracker is run on each sequence and a set of $N$ per-sequence average overlaps $\{ M_j \}_{j=1:N}$ is calculated. The final performance is reported as the average over the sequences, i.e., an overall average overlap $M={1 \over N}\sum\nolimits_{j=1:N} M_j$. The aim of the estimator (evaluation methodology) is to recover the hidden average performance $\mu_A$. In the following we will study the expected value and the variance of the output $M$ depending on whether the tracker is re-initialized at failure (WIR) or the failure is ignored (NOR).

The NOR-based methodologies (\cite{Wu2013,Smeulders2013}) do not detect the failures and the overlaps after the failure affect the estimate of the true overlap $\mu_A$. Alternatively, the WIR-based methodology (our approach) detects a failure, skips $\Delta$ frames and re-initializes the tracker. It can be shown that the expected value $\langle M_\mathrm{NOR} \rangle$ and the variance $var(M_\mathrm{NOR})$ of the overall overlap $M_\mathrm{NOR}$ estimated without re-initialization on the theoretical tracking experiment are
\begin{eqnarray}\label{eq:otb_est_overall}
\langle M_\mathrm{NOR} \rangle = \mu_A(1-\frac{p}{2}), \\
\mathrm{var}(M_\mathrm{NOR}) = \frac{(2-p)\sigma_A^2 }{2 N N_\mathrm{s}} + \frac{p(4-3p)\mu_A^2}{12 N },
\end{eqnarray}
while the expected values and variance for the overall overlap estimated by WIR, i.e., $M_\mathrm{WIR}$, are
\begin{eqnarray}\label{eq:vot_est_overall}
\langle M_\mathrm{WIR} \rangle = \mu_A, \\
\mathrm{var}(M_\mathrm{WIR}) = \sigma_A^2 \frac{N_\mathrm{s}-\Delta(1-p)}{N N_\mathrm{s}(N_\mathrm{s}-\Delta)} \le \mathrm{var}(M_\mathrm{NOR}).\label{eq:vot_est_overall_last}
\end{eqnarray}
Please see the outline of derivation in Appendix~\ref{app:DeriveNORWIR}.

The following observations can be deduced from Eqs. (\ref{eq:otb_est_overall}-\ref{eq:vot_est_overall_last}). The NOR estimator is biased increasingly with the probability of failing at a critical point. If critical points always cause a failure, i.e., $p=1$, then the overall average estimated by the NOR is half the true overlap. On the other hand, the WIR estimator is unbiased, recovers the true hidden overlap, and the mean does not depend on the critical points. The variance of the NOR estimator depends both on the variance of overlaps during successful track as well as the hidden overlap $\mu_A$. This results in a large variance for trackers that track at high overlap and fail at critical points. On the other hand, the variance of the WIR does not show this effect and is always lower than for NOR, i.e., $ \mathrm{var}(M_\mathrm{WIR}) \le \mathrm{var}(M_\mathrm{NOR})$.

\begin{figure} 
    \centering
    \includegraphics[width =7cm]{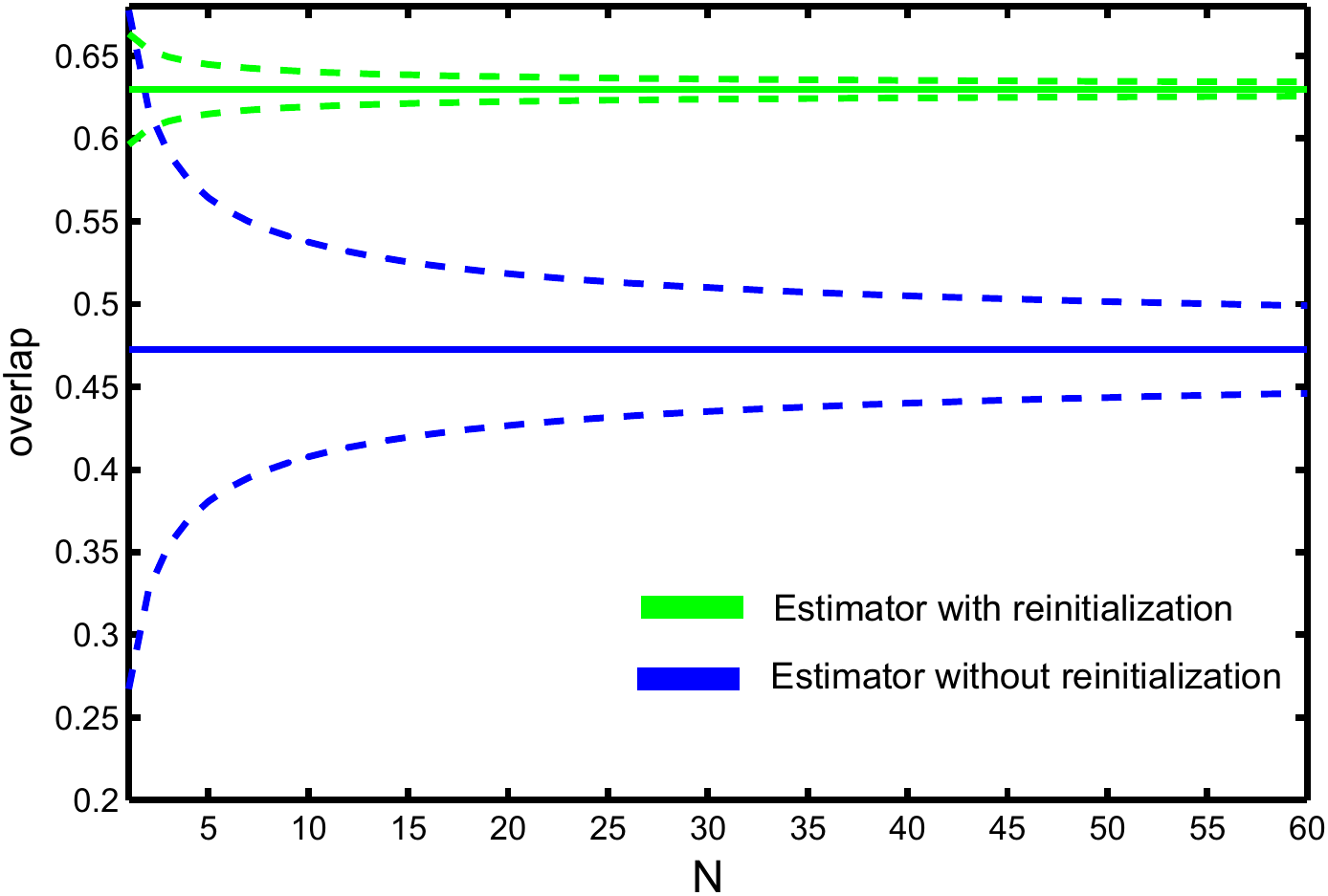}
    \caption{Effects of re-initialization in performance estimators. The expected values and standard deviations of the estimators are shown in solid and dashed lines, respectively.}
    \label{fig:theoret_overall}
\end{figure}

The asymptotic properties of the estimators are visualized in Figure~\ref{fig:theoret_overall} w.r.t. the number of test sequences $N$ for parameters $\mu_A=0.63$, $\sigma_A=0.4$, $N_\mathrm{s}=150$, $p=0.5$, $\Delta=15$. Note that the WIR estimator is indeed asymptotically unbiased, while the NOR is biased toward a lower overlap values. Furthermore, the variance of the WIR is significantly smaller than that of NOR and decreases faster than for WIR, which is primarily due to the second term in $\mathrm{var}(M_\mathrm{NOR})$ (\ref{eq:otb_est_overall}), i.e., lack of re-initializations in NOR. A practical implication is that the methodologies like~\cite{Wu2013,Smeulders2013} require many more sequences than our methodology to produce similarly small variance of the estimate and their estimate will always be much more biased than ours when failures occur. Note that our theoretical model assumes sequences of equal length. If this constrained was further relaxed such that some sequences were allowed to be significantly longer than the others, it would not affect the WIR estimator, but would significantly increase the variance of the NOR even further.

\subsubsection{The importance of per-frame annotation}\label{sec:TheoryPerFrameAnnotation}

To study the impact of visual property annotation strategies, we will assume running a tracker on a dataset in which $N$ sequences contain a particular attribute, e.g., a illumination change. The aim is to estimate tracking performance on this visual attribute. A tracker is thus run on each of $N$ sequences, recovering the set of per-sequence overlaps $\{ M_j \}_{j=1:N}$, and the average of these is reported as an overall performance, i.e., $M={1 \over N}\sum\nolimits_{j=1:N} M_j$.
For ease of exposition assume that each sequence contains $N_\mathrm{A}$ frames with illumination change and the remaining $N_\mathrm{B}=\eta N_\mathrm{A}$ frames contain the other attributes. Thus the per-frame overlaps during the $N_\mathrm{A}$ frames can be described as samples from a distribution with mean $\mu_\mathrm{A}$ and variance $\sigma_A^2$, while the per-frame overlaps in the remaining $N_\mathrm{B}$ frames are governed by a distribution with mean $\mu_\mathrm{B}$ and variance $\sigma_B^2$. For clarity of the analysis we will assume that there are no critical points in any sequence, i.e., a tracker never fails during tracking, and that the variances $\sigma_A^2$ and $\sigma_B^2$ are equal.

A global visual property annotation strategy (GLA) (e.g.,~\cite{Wu2013,Smeulders2013}) calculates overall per-visual property performance $M_\mathrm{GLA}$ using all the frames in sequences that contain at least one frame with the considered visual property. Alternatively, the per-frame annotation strategy (PFA) (our approach) considers only frames annotated with a particular visual attribute to estimate the performance $M_\mathrm{PFA}$. Note, however, that some frames may be incorrectly annotated. From the perspective of bias in state estimation, the most critical frames are those that are incorrectly annotated as the considered attribute. Assume therefore, that in each sequence, a set of $\beta N_\mathrm{A}$ are added as false annotations to the correctly annotated $N_\mathrm{A}$ frames. With these definitions, it is easy to show that the mean and variance of the $M_\mathrm{GLA}$ estimator are
\begin{eqnarray}\label{eq:otb_per_frame}
\langle M_\mathrm{GLA} \rangle = \frac{1}{1+\eta}\mu_A + \frac{\eta}{1+\eta}\mu_B,  \\
\mathrm{var}(M_\mathrm{GLA}) = \frac{1}{N N_\mathrm{A} (1+\eta)}\sigma_A^2,
\end{eqnarray}
while the mean and variance for the and $M_\mathrm{PFA}$ estimator are,
\begin{eqnarray}\label{eq:vot_per_frame}
\langle M_\mathrm{PFA} \rangle = \frac{1}{1+\beta}\mu_A + \frac{\beta}{1+\beta}\mu_B,  \\
\mathrm{var}(M_\mathrm{PFA}) = \frac{1}{N N_\mathrm{A} (1+\beta)}\sigma_A^2.\label{eq:vot_per_frame_last}
\end{eqnarray}

According to equations (\ref{eq:otb_per_frame},\ref{eq:vot_per_frame_last}) both estimators are biased, but the bias in $M_\mathrm{GLA}$ is much greater than the bias in $M_\mathrm{PFA}$. For example, assuming sequence lengths $N_\mathrm{S}=150$, with $N_\mathrm{A}=50$ properly labelled frames and five frames per sequence mislabelled, results in $\eta=2$ and $\beta=0.1$. This means that $M_\mathrm{GLA}$ is biased with $0.67\mu_B$, while the bias of $M_\mathrm{PFA}$ is only $0.09\mu_B$. In fact, since typical sequences contain only small subsets of frames with particular visual attribute, (\ref{eq:otb_per_frame}) shows that the $M_\mathrm{GLA}$ estimator actually reflects performance that is dominated by the other visual attributes, thus significantly skewing the per-visual attribute performance evaluation. Note that the variance of the $M_\mathrm{GLA}$ is lower than that of $M_\mathrm{PFA}$ by a constant $\frac{1+\beta}{1+\eta}$ since it applies more frames. Nevertheless, the variances of both estimators decrease linearly with factor $N N_\mathrm{A}$. A practical implication of these results is that per-frame annotation of moderately-sized dataset (our approach), even with a reasonable number of mislabelled frames, provides a much better estimate of true per-visual attribute performance than a per-sequence labelled large dataset (methodologies used in~\cite{Wu2013,Smeulders2013}).

\section{Experimental evaluation}\label{sec:experiments}

\subsection{VOT2014 challenge}

The tracker comparison methodology from Section~\ref{sec:evalMethodology} was applied to a large-scale experiment, organized as a Visual Object Tracking challenge\footnote{http://www.votchallenge.net/}~(VOT2014). An annotated dataset (Section~\ref{sec:vot_dataset}) was constructed and an evaluation system implemented in Matlab/Octave to fulfill the multi-platform, multi-programming language compatibility requirement from Section~\ref{sec:Requirements}. A minimal API is defined to integrate a tracker with the system regardless of the programming language used to implement the tracker. The reader is referred to the evaluation kit document~\cite{Kristan2014} for further details. Researchers were invited to participate by downloading the evaluation kit, to integrate it into their trackers and to run it locally on their machines. The evaluation kit downloaded the VOT2014 dataset and performed a set of pre-defined experiments (Section~\ref{sec:vot_experiments}). To ensure a fair analysis, the authors were instructed to select a single set of parameters for all experiments. This way, the authors of the trackers themselves were responsible for setting the proper parameters and removing possible errors from the tracker implementations. The raw results from the evaluation system were then submitted to the VOT2014 homepage, along with a short description of the trackers and optionally with the binaries or source code to allow the VOT2014 committee further verification of their results.

\subsubsection{Experiments}\label{sec:vot_experiments}

The VOT2014 challenge includes the following two experiments:
\begin{itemize}
    \item Experiment~1 (\textit{baseline}) runs a tracker on all sequences in the VOT2014 dataset by initializing it on the ground truth bounding boxes.
    \item Experiment~2 (\textit{bounding box perturbation}) performs Experiment~1 with noisy bounding boxes. The noise affected the position and size by drawing perturbations uniformly from the $\pm 10\%$ interval of the ground truth bounding box size and the rotation by drawing uniformly from the $\pm 0.1$ radian range. 
\end{itemize}

All the experiments were automatically performed by the evaluation kit\footnote{\vottoolkitpage}. A tracker was run on each sequence 15 times to obtain a better statistics on its performance.

\subsubsection{Tested trackers}

In total $38$ trackers were considered in the challenge, most of which had been published in recent years and represent the state-of-the-art. These included $33$ original submissions and $5$ baseline highly-cited trackers that were contributed by the VOT committee. We reference the unpublished trackers by the VOT2014 challenge report~\cite{KRISTAN_2014_ECCV}. For the interested readers a more detailed description of each tracker can be found in the supplementary material and a condensed summary of the trackers is available in Table~\ref{tab:ranking}.

Several trackers explicitly decomposed the target into parts. These ranged from key-point-based trackers CMT~\cite{Nebehay2014WACV}, IIVTv2~\cite{KRISTAN_2014_ECCV}, Matrioska~\cite{Maresca2013} and its derivative MatFlow~(a combination of Matrioska and FoT~\cite{vojir2011cvww}) to general part-based trackers LT-FLO~\cite{Lebeda2013vot}, PT+~(an improvement of the Pixeltrack tracking algorithm~\cite{2013_DUFFNER}), LGT~\cite{Cehovin2013}, OGT~\cite{2014_NAM}, DGT~\cite{2014_CAI}, ABS~\cite{KRISTAN_2014_ECCV}, while three trackers applied flock-of-trackers approaches FoT~\cite{vojir2011cvww}, BDF~\cite{2014_MARESCA_a} and FRT~\cite{Adam2006}. Several approaches were applying global generative visual models for target localization: a channel blurring approach EDFT~\cite{Felsberg2013vot} and its derivative qwsEDFT~\cite{2014_FELSBERG}~(an improvement of both trackers DFT~\cite{sevilla2012cvpr} and EDFT~\cite{Felsberg2013vot}), GMM-based VTDMG~(an extension of~\cite{2012_YIJEONG}), scale-adaptive mean shift eASMS~(an extension of ASMS~\cite{2014_VOJIR}), color and texture-based ACAT~(a combination of Colour Attributes Tracker~(CAT)~\cite{2014_DANELLJAN} and CSK tracker~\cite{2012_HENRIQUES}), NCC based tracker with motion model IMPNCC~(an improvement of the NCC tracker~\cite{2001_BRIECHLE}), two color-based particle filters SIR-PF~(a combination of particle filter, a background model as in~\cite{comaniciuTPAMI2003} and information coming from colour space~YCbCr) and IPRT~(an improvement of colour-based particle filter~\cite{Nummiaro03,Perez02} using particle re-propagation), a compressive tracker CT~\cite{zhang2012eccv} and intensitiy-template-based pca tracker IVT~\cite{Ross2008}. Two trackers applied fusion of flock-of-trackers and mean shift, HMM-TxD~\cite{KRISTAN_2014_ECCV} and DynMS~(which is a Mean Shift tracker~\cite{2002_COMANICIU} with an isotropic kernel bootstrapped by a flock-of-features (FoF) tracker). Many trackers were based on discriminative models, i.e., boosting-based particle filter MCT~\cite{2014_DUFFNER}, multiple-instance-learning-based tracker MIL~\cite{babenko2011tpami}, detection-based FSDT~\cite{KRISTAN_2014_ECCV} while several applied regression-based techniques, i.e., variations of online structured SVM, Struck~\cite{hare2011iccv}, aStruck~(a combination of optical-flow-based tracker and the discriminative tracker Struck~\cite{hare2011iccv}), TStruck~(a CUDA-based implementation of the Struck tracker~\cite{hare2011iccv}), $\mathrm{PLT}_{13}$~\cite{Kristan2013a} and $\mathrm{PLT}_{14}$~(an improved version of $\mathrm{PLT}_{13}$ tracker), kernelized-correlation-filter-based KCF~\cite{Henriques2014}, kernelized-least-squares-based ACT~\cite{2014_DANELLJAN} and discriminative correlation-based DSST~\cite{2014_DANELLJAN_BMVC} and SAMF~\cite{2014_LIZHU}.

\subsection{The VOT2014 Dataset}\label{sec:vot_dataset}

A usual approach to creating a diverse dataset is collecting all sequences from existing datasets. However, a large dataset does not necessarily mean being rich in visual properties. In fact, many sequences may be visually similar and would not contribute to the diversity while they would significantly slow down the evaluation process. We have therefore applied an approach that leads to a dataset that includes various visual phenomena while containing a small number of sequences.

The dataset was prepared as follows. The initial pool included $394$ sequences, including sequences used by various authors in the tracking community, the VOT2013 benchmark~\cite{Kristan2013a}, the recently published ALOV dataset~\cite{Smeulders2013}, the Online Object Tracking Benchmark~\cite{Wu2013} and additional, so far unpublished, sequences. The set was manually filtered by removing sequences shorter than $200$ frames, grayscale sequences, sequences containing poorly defined targets (e.g., fireworks) and sequences containing cuts. The following global intensity (it) and spatial (sp) attributes were automatically computed for each of the $193$ remaining sequences:
\begin{enumerate}
    \item \textit{Illumination change} is defined as the average of the absolute differences between the object intensity in the first and remaining frames (it).
    \item \textit{Object size change} is the sum of averaged local size changes, where the local size change at frame $t$ is defined as the average of absolute differences between the bounding box area in frame $t$ and past fifteen frames (sp).
    \item \textit{Object motion} is the average of absolute differences between ground truth center positions in consecutive frames (sp).
    \item \textit{Clutter} is the average of per-frame distances between two histograms: one extracted from within the ground truth bounding box and one from an enlarged area (by factor 1.5) outside of the bounding box (it).
    \item \textit{Camera motion} is defined as the average of translation vector lengths estimated by key-point-based RANSAC between consecutive frames (sp).
    \item \textit{Blur} was measured by the Bayes-spectral-entropy camera focus measure~\cite{Kristan2005a} (it).
    \item \textit{Aspect-ratio change} is defined as the average of per-frame aspect ratio changes. The aspect ratio change at frame $t$ is calculated as the ratio of the bounding box width and height in frame $t$ divided by the ratio of the bounding box width and height in the first frame (sp);
    \item \textit{Object color change} defined as the change of the average hue value inside the bounding box (it);
    \item  \textit{Deformation} is calculated by dividing the images into $8$~$\times$~$8$ grid of cells and computing the sum of squared differences of averaged pixel intensity over the cells in current and first frame (it).
    \item \textit{Scene complexity} represents the level of randomness (entropy) in the frames and it was calculated as $e = \sum_{i = 0}^{255}b_i\log b_i$, where $b_i$ is the number of pixels with value equal to $i$ (it).
\end{enumerate}
In this way each sequence was represented as a $10$-dimensional feature vector. Sequences were clustered in an unsupervised way using affinity propagation~\cite{Frey2007} into $12$ clusters\footnote{The parameters were automatically set. We checked that small perturbations did not result in different clusterings.}. From these, $25$ sequences were manually selected such that the various visual phenomena like, occlusion, were still represented well within the selection.

The selected objects in each sequence are manually annotated by bounding boxes. For most sequences, the authors provide axis-aligned bounding boxes placed over the target. For most frames, the axis-aligned bounding boxes approximated the target well with large percentage of pixels within the bounding box (at least $>60\%$) belonging to the target. Some sequences contained elongated, rotating or deforming targets and these were re-annotated by rotated bounding boxes. After inspecting all the bounding box annotations, sequences with misplaced original annotations were re-annotated.

\begin{figure} 
    \centering
    \includegraphics[width=8.5cm]{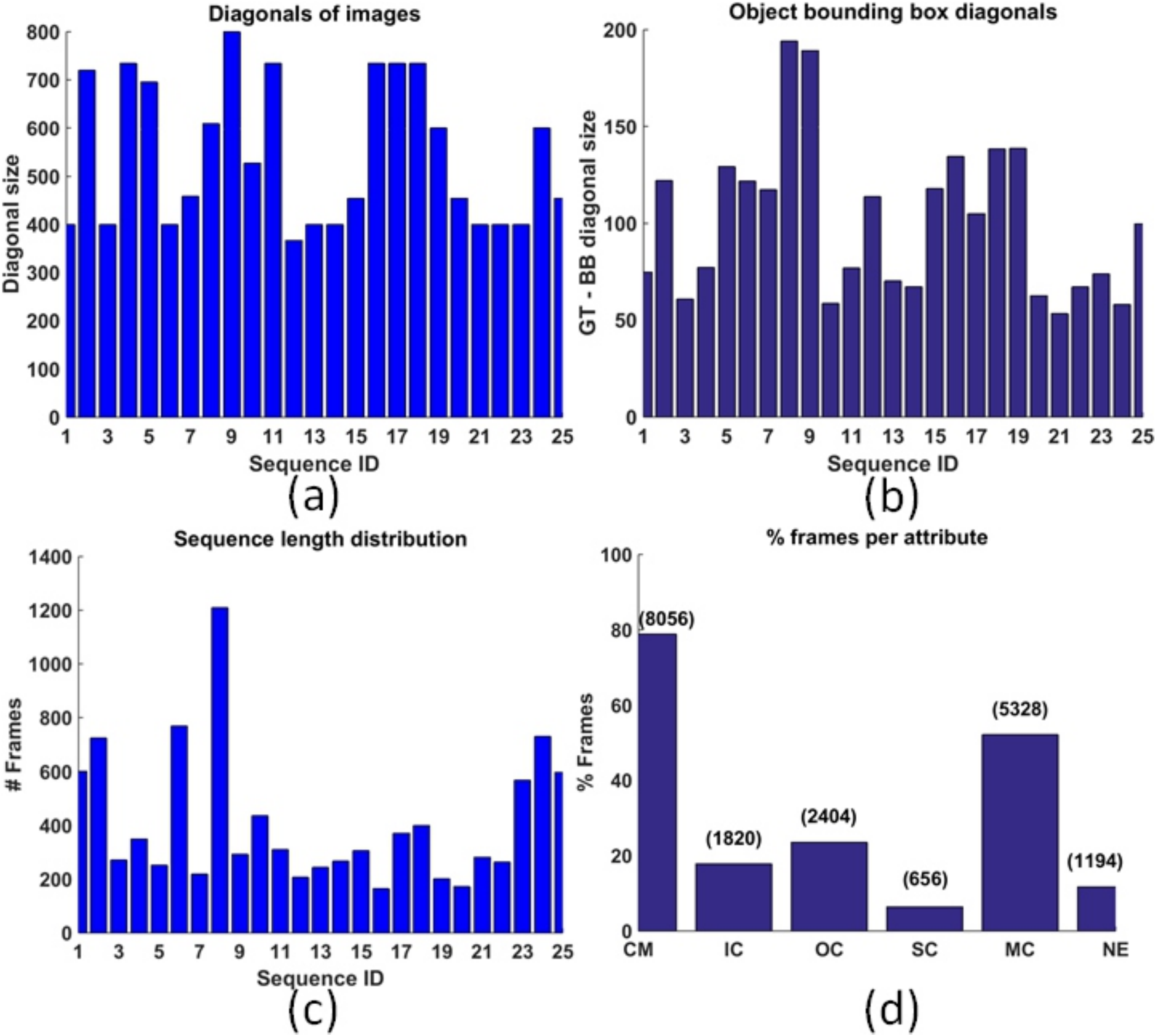}
    \caption{Summary of the VOT2014 dataset properties: frames sizes per sequence (a), ground truth bounding box sizes per sequence (b), number of frames per sequence (c), percentage of frames per visual attribute with number of frames per attribute in parentheses (d). The abbreviations CM, IC, OC, SC, MC and NE stand for camera motion, illumination change, occlusion, scale change, motion change and neutral attributes, respectively.}
    \label{fig:datasetstats}
\end{figure}

Additionally, we labeled each frame in each sequence with five visual attributes that reflect a particular challenge in appearance degradation: (1)~camera motion, (2)~illumination change, (3)~motion change, (4)~size change and (5)~occlusion. In case a particular frame  had none of the five attributes, we labeled the frame  as (6)~neutral. A summary of sequence properties is presented in Figure~\ref{fig:datasetstats}. The average length of consecutive frames containing an attribute was $335.6$ for camera motion, $107.1$ for illumination change, $16.9$ for occlusion, $27.7$ for motion change, $34.5$ for occlusion, and $99.5$ for neutral frames.

\subsubsection{Estimation of practical difference thresholds}\label{sec:est_of_prac_diff}

The practical difference (Section~\ref{sec:prac_difference}) strongly depends on the target as well as the free parameters of the annotation model. Ideally, a per-frame estimate of $\gamma$ would be required for each sequence, but that would present a significant undertaking. On the other hand, using a single threshold for the entire dataset is too restrictive as the properties of targets vary across the sequences. A compromise can be taken in this case by computing a single threshold per sequence. We propose selecting $M$ frames per sequence and have $J$ expert annotators place the bounding boxes carefully $K$ times on each frame. In this way $N=K\times J$ bounding boxes are obtained per frame. One of the bounding boxes can be taken as a possible ground truth and $N-1$ overlaps can be computed with the remaining ones. Since all annotations are considered ``correct'', any two overlaps should be considered equivalent, therefore the difference between these two overlaps is an example of negligibly small difference. By choosing each of the bounding boxes as ground truth, $M(N((N - 1)^2 - N + 1))/2$ samples of differences are obtained per sequence. The practical difference threshold per sequence is estimated as the average of these values.

\begin{figure}
        \centering
        \includegraphics[width=0.45\textwidth]{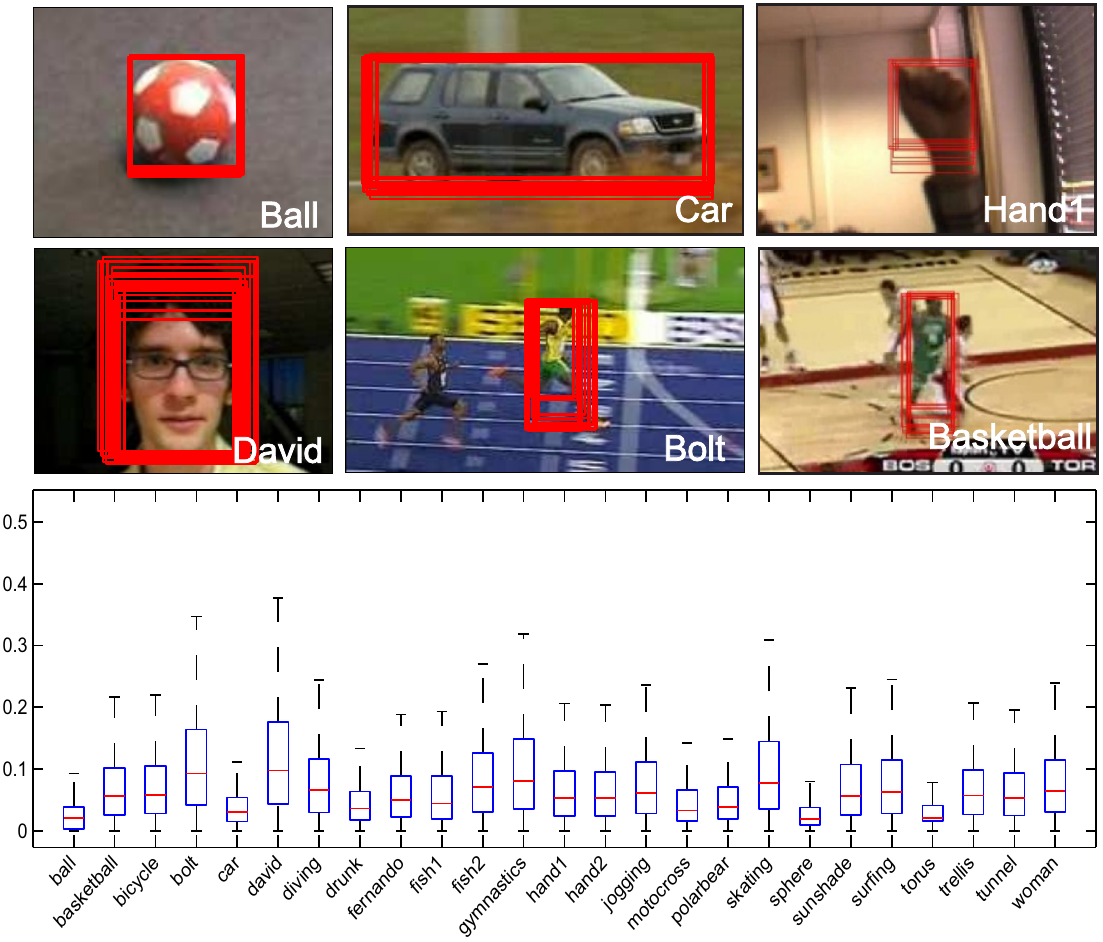}
        \caption{\label{fig:annotExamples} Examples of the diversity of bounding box annotations for different images (top) and box plots of per-sequence distribution of ground truth overlaps.}
\end{figure}

Seven experts have annotated four frames per sequence three times. A single frame with an overlayed ground truth bounding box per sequence was displayed during annotation, serving as a guideline of what should be annotated. Thus a set of 15960 samples of differences was obtained per sequence and used to compute the per-sequence practical difference thresholds. The boxplots of the differences are shown in Figure~\ref{fig:annotExamples} along with a few frames with overlaid annotations. It is clear that the threshold on practical difference varies over the sequences. For the sequences containing rigid objects, the practical difference threshold is small (e.g., ball), but becomes large for sequences with deformable/articulated objects (e.g., bolt).

\subsection{Study of the methodology parameters}

\subsubsection{Estimation of the burn-in period}\label{sec:burnin_exp}  

\begin{figure} 
    \centering
    \includegraphics[width=8.5cm]{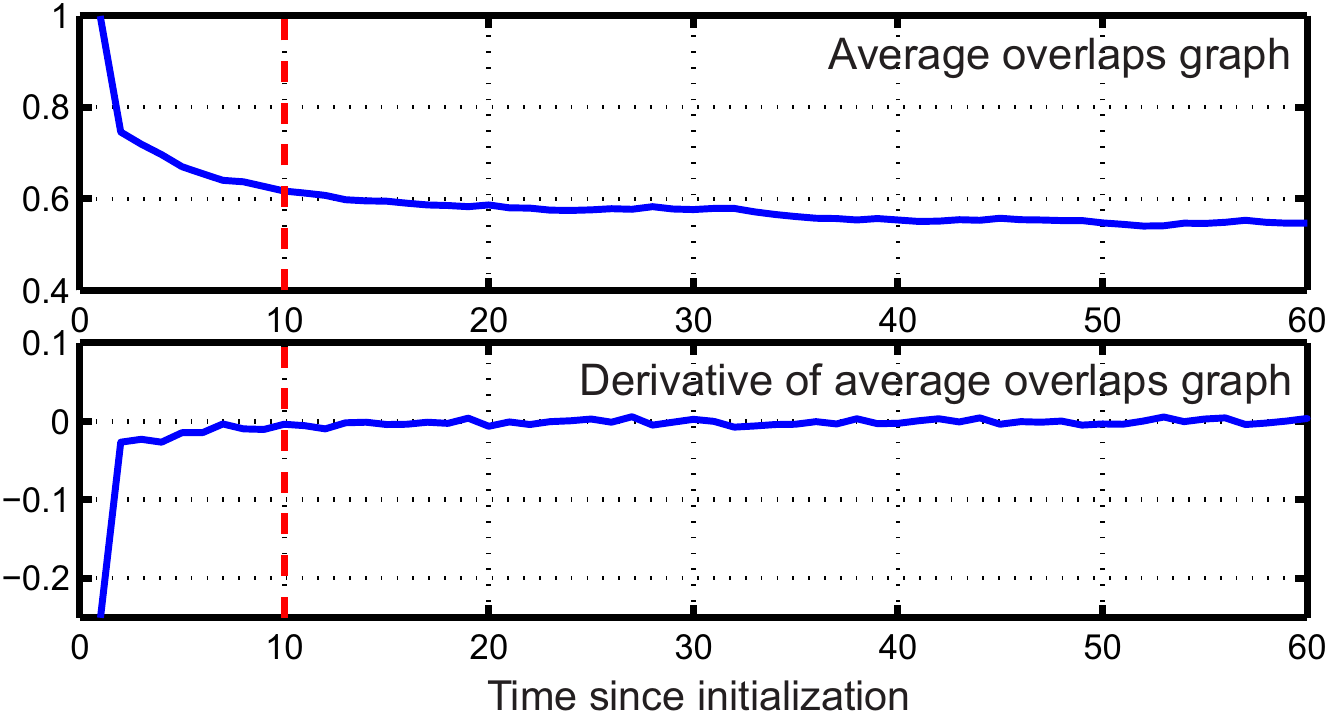}
    \caption{Overlaps after re-initialization averaged over a large number of trackers and many re-initializations (top) and the derivative of this graph with respect to time (bottom). The derivative becomes negligible after 10 frames.}
    \label{fig:burnIn}
\end{figure}

A study was designed to estimate the burn-in period. Seven trackers were run with re-initialization on the VOT2013 dataset~\cite{KristanVOT2013}, which contains sequences recorded at approximately 20 frames per second. After each re-initialization we recorded the per-frame overlaps with the ground truth (an overlap sequence). Using this protocol we obtained 3249 overlap sequences, which were averaged into a single average overlap sequence shown in Figure~\ref{fig:burnIn}. The rate of temporal change in overlap is characterized by the derivative of this sequence (also shown in Figure~\ref{fig:burnIn}). It is apparent that the rate of overlap change stabilizes at ten frames after re-initialization. We have therefore set the burn-in period to $N_\mathrm{burnin}=10$ frames in our methodology.

The effect of the burn-in period was further quantified by running several state-of-the-art trackers STRUCK~\cite{hare2011iccv}, DSST~\cite{2014_DANELLJAN_BMVC}, SAMF~(\cite{2014_LIZHU}) and KCF~\cite{Henriques2014} and two trackers commonly used as baselines, CT~\cite{zhang2012eccv} and FRT~\cite{Adam2006} on the VOT2014 dataset. Table~\ref{tbl:results_burnin} shows the average accuracy for different values of the burn-in period. The average accuracy is, as expected, slightly reduced when including the frames from the burn-in period. The extent of the drop in accuracy is larger for trackers that fail more often.

\begin{table}\caption{Influence of different burn-in values on raw accuracy.
\label{tbl:results_burnin}}
\resizebox{\columnwidth}{!}{ 
\begin{tabular}{c||c c c c c c|c}
$N_\mathrm{burnin}$ & \textbf{DSST} & \textbf{KCF} & \textbf{SAMF} & \textbf{CT} & \textbf{FRT} & \textbf{Struck} & \textbf{Average} \\\hline
\textbf{0} & 0.6293 & 0.6386 & 0.6213 & 0.4273 & 0.4871 & 0.5167 & 0.5534 \\
\textbf{2} & 0.6285 & 0.6378 & 0.6205 & 0.4248 & 0.4838 & 0.5143 & 0.5516 \\
\textbf{4} & 0.6273 & 0.6369 & 0.6195 & 0.4209 & 0.4786 & 0.5103 & 0.5489 \\
\textbf{6} & 0.6264 & 0.6370 & 0.6191 & 0.4183 & 0.4749 & 0.5071 & 0.5471 \\
\textbf{8} & 0.6258 & 0.6376 & 0.6192 & 0.4165 & 0.4726 & 0.5047 & 0.5461 \\
\textbf{10} & 0.6256 & 0.6385 & 0.6198 & 0.4149 & 0.4711 & 0.5029 & 0.5455 \\
\end{tabular}
}
\end{table}

\subsubsection{Influence of the re-initialization frame skipping}\label{sec:skipping_exp}

As explained in Section~\ref{sec:vot_eval_met}, $N_\mathrm{skip}$ frames are skipped after re-initialization to remove the bias of potentially re-initializing the tracker on the same visual content that caused the failure. The effect of the $N_\mathrm{skip}$ values was quantified by re-running the trackers from previous section on the VOT2014 dataset. The number of failures and robustness ranks w.r.t. the skipping values  $N_\mathrm{skip}$ are shown in Table~\ref{tab:results_skip}. The number of failures most significantly changes between one to three skipped frames and remains stable with increasing $N_\mathrm{skip}$. The relative changes are consistent across trackers. This is confirmed by the ranking, which remains stable. 

\begin{table}\caption{Robustness raw and rank values for different values of frames skipped $N_\mathrm{skip}$.
\label{tab:results_skip}}
\resizebox{\columnwidth}{!}{ 
 \begin{tabular}{ c c || r r r r r r }
  $N_\mathrm{skip}$ &  R  &  DSST  &  KCF  &  SAMF &  CT & FRT & Struck \\ \hline 
	1 & raw & 1.32 & 2.04 & 1.44 & 1.93 & 3.76 & 3.28 \\
	3 & raw & 1.12 & 1.84 & 1.56 & 1.90 & 3.68 & 2.76 \\ 
	5 & raw & 1.16 & 1.44 & 1.36 & 1.57 & 3.36 & 2.72 \\
	7 & raw & 1.16 & 1.56 & 1.36 & 1.55 & 3.48 & 2.36 \\
	9 & raw & 1.00 & 1.52 & 1.16 & 1.54 & 2.96 & 2.28 \\
\hline
	1 & rank & 2.58 & 3.32 & 2.74 & 3.40 & 5.06 & 3.86 \\
	3 & rank & 2.44 & 3.18 & 3.02 & 3.28 & 5.26 & 3.82 \\
	5 & rank & 2.64 & 3.02 & 2.94 & 3.34 & 5.16 & 3.90 \\
	7 & rank & 2.70 & 3.12 & 2.86 & 3.20 & 5.38 & 3.74 \\
	9 & rank & 2.60 & 3.32 & 2.82 & 3.36 & 4.94 & 3.96 \\
 \end{tabular}
}
\end{table}

\subsubsection{Influence of difference tests} 

The proposed methodology applies tests of performance equivalence by testing statistical and practical differences in tracker performance. In absence of these tests, trackers that perform slightly differently in average values of performance measures would be assigned different ranks even tough the difference in performance might not be statistically significant or below the annotation noise level (practical difference). To quantify the variations in ranks, we sampled 50 random sub-sets of 15 sequences from VOT2014 dataset, ranked DSST, KCF, SAMF, CT, FRT and Struck on all subsets and computed the average of the rank variances over all trackers.
Table~\ref{tab:rank_var} reports the rank variations for sequence-pooled and attribute-normalized ranking. The difference tests consistently reduce the variance in both setups.

\begin{table}\caption{Rank variance (var.) with (T) and without (N) difference tests for accuracy   and robustness computed for sequence-pooled (Seq. pool.) and attribute-normalized (Att. norm.) setting.\label{tab:rank_var}}
\centering
 \begin{tabular}{ l ||c c | c c || c c  | c c }
   & \multicolumn{4}{ c}{accuracy} & \multicolumn{ 4 }{c}{robustness} \\
   & \multicolumn{2}{ c |}{Seq. pool.} & \multicolumn{2}{c ||}{Att. norm.} & \multicolumn{2}{ c |}{Seq. pool.} & \multicolumn{2}{ c }{Att. norm.} \\
   & T & N & T & N& T & N & T & N\\
   \hline
   var & 0.1 & 0.11 & 0.26 & 0.31  &  0.07 & 0.1 & 0.09 & 0.34 \\
 \end{tabular}
\end{table}

\subsection{Comparison with related methodologies}

Performance evaluation methodologies mainly differ in use of re-initialization and detail of visual attribute annotation in sequences. The theoretical predictions derived in Section~\ref{sec:theoreticalEval} were again validated experimentally on the VOT2014 dataset using the trackers from previous section.
 
\subsubsection{Effects of re-initialization}

The theoretical comparison of estimators (Section~\ref{sec:TheoryReinit}) that apply re-initialization, ($M_\mathrm{WIR}$), and those that do not, $M_\mathrm{NOR}$, was evaluated experimentally. Each tracker was run on all sequences in the VOT2014 dataset once with re-intializations and once without. A set of $K$ sequences was randomly sampled and average overlap was computed on this set for each estimator. The process was repeated thousand times for $K<24$ to estimate the mean and variance. For $K=24$ there are only 25 possible different combinations of sequences, therefore the mean and variance were computed only on these. Table~\ref{tab:resultsReinit} shows results for varying $K$. Due to sampling with replacement, sequences were repeated across the sets, which means that the variance was underestimated, especially for the $K=24$. The actual variances of the average accuracy are expected to be higher. Nevertheless, the relative trends are as predicted by the theoretical model. The means of $M_\mathrm{NOR}$ are consistently lower than for $M_\mathrm{WIR}$, which is especially evident for trackers that fail frequently, e.g., FRT and CT. Moreover, the variance of $M_\mathrm{NOR}$ is consistently higher than for $M_\mathrm{WIR}$ across all trackers. The Wilcoxon paired tests showed that both types of differences are statistically significant at $p<0.01$.

\begin{table}[pt!]\caption{Performance of estimators with re-initialization, $\mathrm{Y}(\mathrm{WIR})$, and without re-initialization, $\mathrm{N}(\mathrm{NOR})$ indicated in the column denoted by R. Average overlap is shown for each tracker and the standard deviation is shown in brackets.}
\label{tab:resultsReinit}
\resizebox{\columnwidth}{!}{
\begin{tabular}{c c||c c c c c c} 
K                  & R & DSST      & KCF       & SAMF      & CT        & FRT       & Struck \\ \hline
\multirow{2}{*}{5} & N   & 0.49(.14) & 0.49(.15) & 0.50(.14) & 0.23(.09) & 0.24(.09) & 0.35(.11) \\ 
                   & Y   & 0.63(.09) & 0.64(.08) & 0.63(.08) & 0.43(.07) & 0.49(.06) & 0.52(.08) \\ \hline
\multirow{2}{*}{10}& N   & 0.50(.10) & 0.49(.10) & 0.51(.10) & 0.24(.06) & 0.24(.06) & 0.36(.08) \\ 
                   & Y   & 0.63(.06) & 0.64(.06) & 0.63(.06) & 0.43(.05) & 0.48(.04) & 0.52(.06) \\ \hline
\multirow{2}{*}{15}& N   & 0.50(.08) & 0.49(.08) & 0.51(.08) & 0.24(.05) & 0.25(.05) & 0.36(.06) \\ 
                   & Y   & 0.63(.05) & 0.64(.05) & 0.63(.05) & 0.43(.04) & 0.49(.04) & 0.52(.05) \\ \hline
\multirow{2}{*}{20}& N   & 0.50(.07) & 0.50(.07) & 0.52(.07) & 0.24(.04) & 0.24(.04) & 0.36(.06) \\ 
                   & Y   & 0.63(.04) & 0.64(.04) & 0.63(.04) & 0.43(.03) & 0.49(.03) & 0.52(.04) \\ \hline
\multirow{2}{*}{24}& N   & 0.50(.06) & 0.50(.07) & 0.52(.06) & 0.24(.04) & 0.24(.04) & 0.36(.05) \\ 
                   & Y   & 0.63(.04) & 0.64(.04) & 0.63(.04) & 0.43(.03) & 0.49(.03) & 0.52(.04) \\ 
\end{tabular}
}
\end{table} 
\subsubsection{Importance of per-frame annotation}

\begin{figure}[h]
    \centering
    \includegraphics[width=8.5cm]{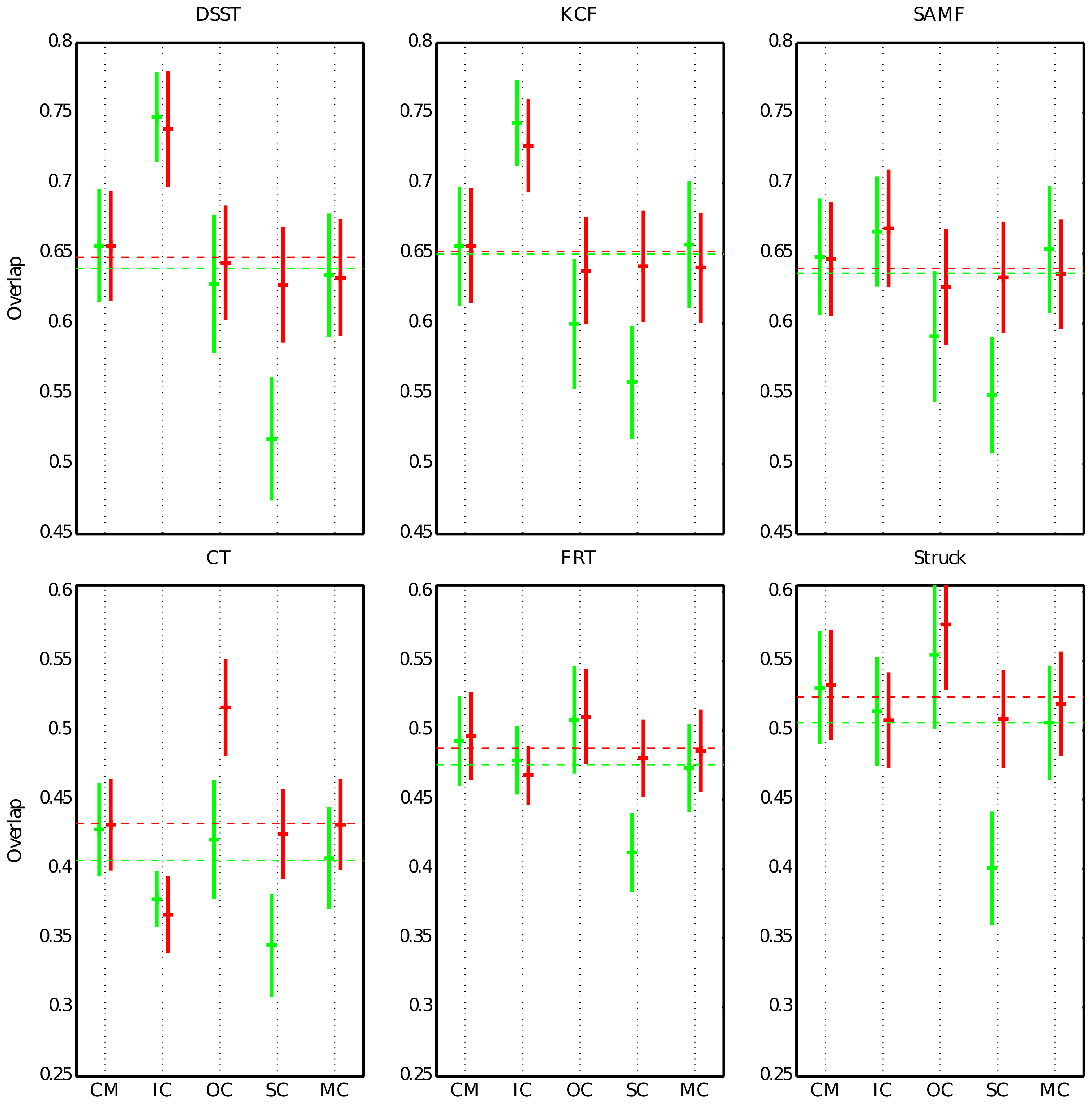}
    \caption{The mean and variance of estimators that apply per-frame (green) and per-sequence (red) visual attribute annotation. The dashed lines show average performance on the dataset. The abbreviations CM, IC, OC, SC, and MC are used for camera motion, illumination change, occlusion, scale change and motion change, respectively.}
    \label{fig:perFrameAnnot}
\end{figure}

The properties of estimators that apply per-frame visual attribute annotation, $M_\mathrm{GLA}$, and the estimators that apply only per-sequence annotation, $M_\mathrm{PFA}$, were estimated using a similar experiment as in previous section. For a fair comparison, re-initialization was applied in all experiments. The results for $K=24$ sequences are visualized in Figure~\ref{fig:perFrameAnnot} and confirm the predictions from our theoretical model. The variance of per-attribute $M_\mathrm{GLA}$ is generally slightly smaller than $M_\mathrm{PFA}$ since $M_\mathrm{GLA}$ uses more frames in estimation, of which many might not contain the attribute in question, making the $M_\mathrm{GLA}$ estimator strongly biased toward the global mean. This bias is also reflected in the dispersion of per-attribute values around their global mean, which is greater for $M_\mathrm{PFA}$ than for $M_\mathrm{GLA}$. This means that the $M_\mathrm{GLA}$ is much weaker at making predictions regarding per-visual attribute performance evaluation. For example, consider the trackers DSST, KCF and SAMF. These are highly similar trackers by design, which is reflected in the trends of per-attribute values in Figure~\ref{fig:perFrameAnnot}. Nevertheless, the $M_\mathrm{GLA}$ cannot distinguish performance with respect to attributes motion change, scale change and occlusion, while the performance difference is clear from $M_\mathrm{PFA}$. A Wilcoxon paired test on pairs with varying $K=15:24$ showed that the variance of $M_\mathrm{PFA}$ is lower than that of $M_\mathrm{GLA}$ at level $p<0.01$ and an F-test on dispersion showed a difference at significance $p<0.05$.
 
\subsection{Application to tracker analysis on VOT2014}

\begin{figure}[h!]
    \centering
    \includegraphics[width=8cm]{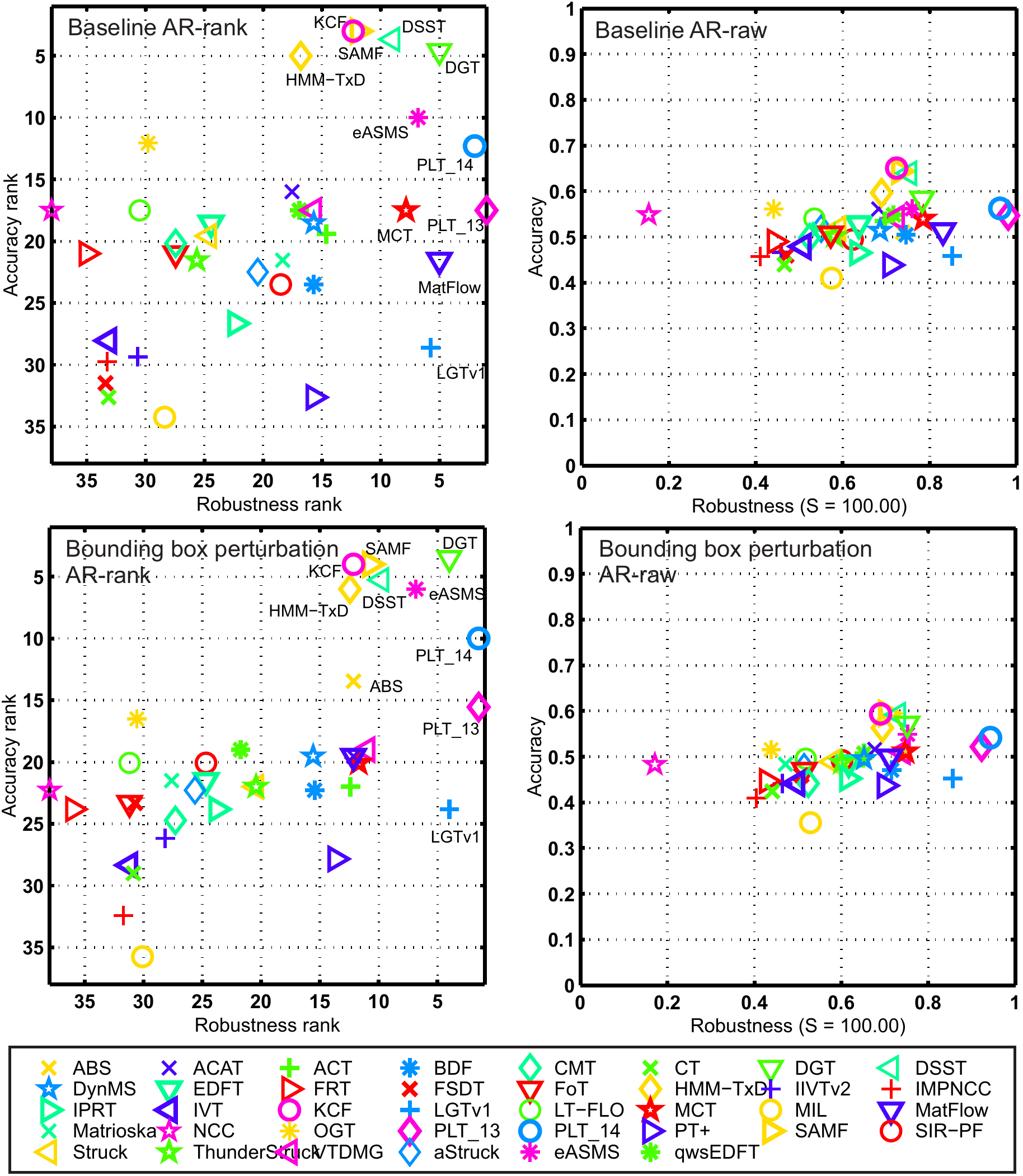}
    \caption{The AR ranking and raw plots for the baseline and bounding box perturbation experiments calculated by sequence-pooled ranking. A tracker is among top-performing if it resides close to the top-right corner of the plot.}
    \label{fig:rankingplots}
\end{figure}

The results of the \textit{baseline} and \textit{bounding box perturbation} experiments described in Section~\ref{sec:vot_experiments} are visualized in Figure~\ref{fig:rankingplots} and summarized in Table~\ref{tab:ranking}. The AR-rank plots in Figure~\ref{fig:rankingplots} are obtained by concatenating results of all sequences into a super-sequence, calculate the average performance measures and calculate the ranks from these. In Table~\ref{tab:ranking}, these results are denoted as sequence-pooled ranking. In addition to rank plots, we show the accuracy/robustness raw plots (AR-raw) as proposed in~\cite{CehovinTracMeas2015} as well. Note that the AR-raw plots~\cite{CehovinTracMeas2015} compute the robustness as the probability of a tracker still tracking after $S$ frames. This parameter affects only scaling, but does not change the order of trackers. We chose $S=100$ to fully utilize the horizontal space in the AR-raw plots.
 
The top-performing trackers in robustness considering both the baseline and noise experiments are $\mathrm{PLT}_{13}$, $\mathrm{PLT}_{14}$, MatFlow and DGT. $\mathrm{PLT}_{13}$ and $\mathrm{PLT}_{14}$ are trackers that apply holistic models. Both trackers are extensions of the Struck~\cite{hare2011iccv} tracker which uses a structured SVM on grayscale patches to learn a regression from intensity to center of object displacement. In contrast to Struck, the $\mathrm{PLT}_{13}$ and $\mathrm{PLT}_{14}$ also apply histogram backprojection as feature selection strategy in the SVM training. The $\mathrm{PLT}_{13}$ is the winner of the VOT2013 challenge~\cite{Kristan2013a} which does not adapt the target size, while the $\mathrm{PLT}_{14}$ is an extension of $\mathrm{PLT}_{13}$ that adapts the size as well. Interestingly, the $\mathrm{PLT}_{14}$ does improve in accuracy compared to $\mathrm{PLT}_{13}$, at a cost of slightly decreased robustness. The MatFlow and DGT are part-based trackers. The MatFlow tracker is an extension of Matrioska~\cite{Maresca2013} which applies a ORB/SURF keypoints and robust voting and matching techniques. Looking at the noise AR-rank plots in Figure~\ref{fig:rankingplots} we see that the rank of Matflow significantly drops compared to PLT trackers. The DGT tracker decomposes a target into parts by superpixels and casts tracking as graph matching between corresponding superpixels across consecutive frames. The DGT also applies segmentation to improve part selection.

In terms of accuracy, the top-performing trackers are DSST, SAMF, KCF and DGT. The DSST, SAMF and KCF are  correlation-filter-based trackers derived from MOSSE~\cite{Bolme2010} that apply holistic models, i.e., a HOG~\cite{Dalal05}. In fact, DSST and SAMF are extensions of the KCF tracker. The similarity in design is reflected in the AR plots (e.g., Figure~\ref{fig:rankingplots}). Note that these trackers form a cluster in the AR-rank space.
 
It is interesting to further study trackers that apply similar concepts for target localization. MatFlow extends Matrioska by applying a flock-of-trackers variant BDF. At a comparable accuracy ranks, the MatFlow by far outperforms the original Matrioska in robustness. The boost in robustness ranks might be attributed to addition of BDF, which is supported by the fact that BDF alone outperforms in robustness the flock-of-trackers tracker FoT as well as trackers based on variations of FoT, i.e., aStruck, HMM-TxD and dynMS. This speaks of resiliency to outliers in flock selection in BDF.

Two trackers combine color-based mean shift with flow, i.e., dynMS and HMM-TxD and obtain comparable ranks in robustness, however, the HMM-TxD achieves a significantly higher accuracy rank, which might be due to considerably more sophisticated tracker merging scheme in HMM-TxD. Both methods are outperformed in robustness by the scale-adaptive mean shift eASMS that applies motion prediction and colour space selection. 

The set of evaluated trackers included the original Struck and two variations, TStruck and aStruck. TStruck is a CUDA-speeded-up TStruck and performs quite similarly to the original Struck in baseline and noise experiment. The aStruck applies the flock-of-trackers for scale adaptation in Struck and improves in robustness on the baseline experiment, but is ranked lower in the noise experiment. This implies that estimation of fewer parameters in Struck results in more accurate and robust performance in cases of poor initialization. This is consistent with the results of comparison of PLT trackers, which are derived from Struck. Note that these trackers by far outperform Struck, which further supports the importance of feature selection in PLT trackers.

\begin{figure}[t!]
    \centering
    \includegraphics[width=8.5cm]{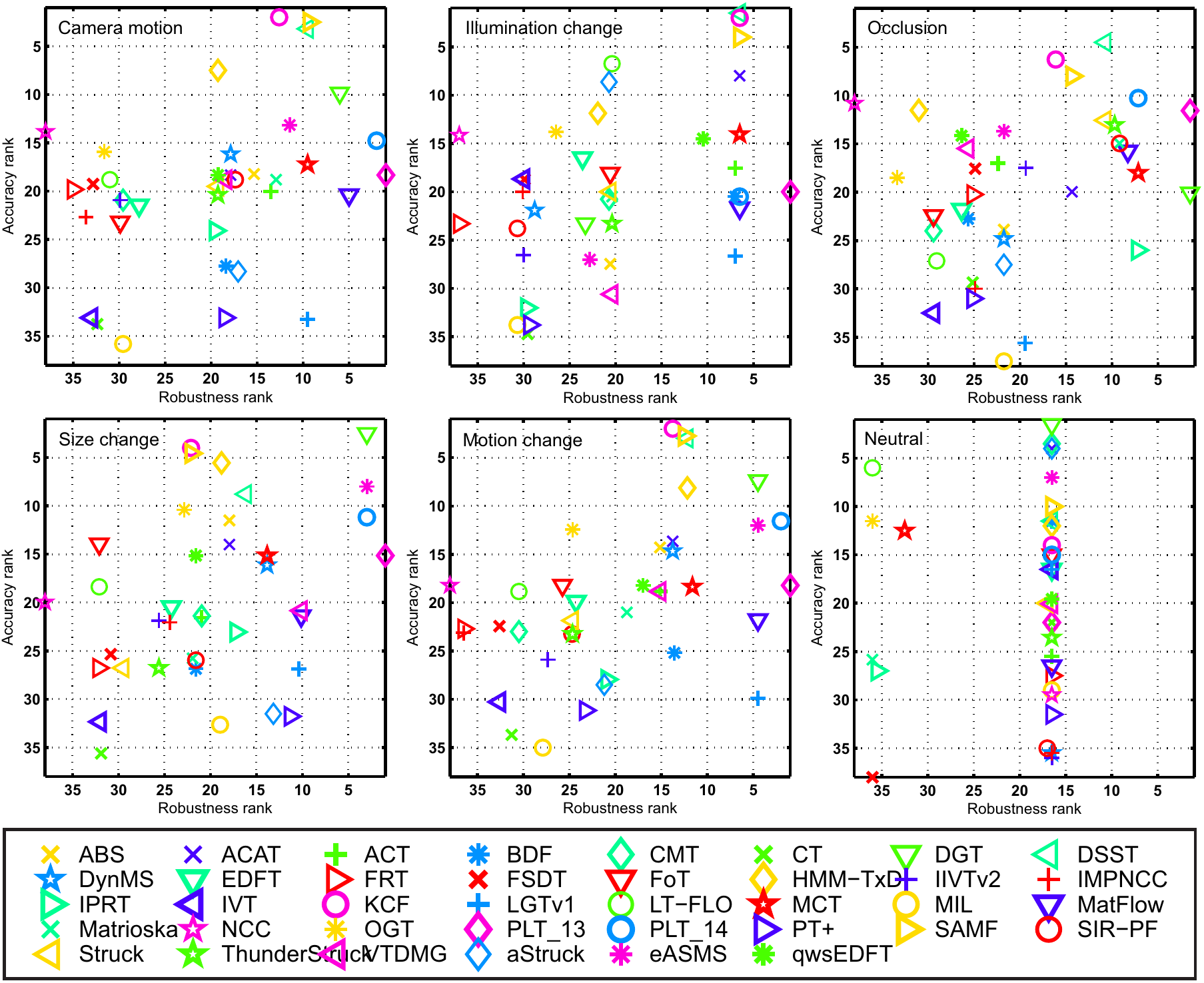}
    \caption{The AR-rank plots of the baseline experiment with respect to the six sequence attributes. A tracker is among top-performing if it resides close to the top-right corner of the plot. }
    \label{fig:rankingplots2}
\end{figure}

The per-visual attribute AR-rank plots are shown in Figure~\ref{fig:rankingplots2}. At illumination changes, trackers form several equivalent classes of robustness. The top-performing trackers in accuracy and robustness remain the DSST, KCF, SAMF and most robust is $\mathrm{PLT}_{13}$. However, the DGT drops drastically in accuracy as well as in robustness, since DGT relies heavily on the color information in segmentation. A similar degradation is observed for the size-adaptive color mean-shift eASMS whose performance also significantly drops a illumination change. Still, the color segmentation in DGT significantly improves tracking during occlusion. The benefits of size adaptation in DGT and eASMS are most apparent from the ranks at size-change and motion-change attributes. The neutral visual attribute does not present particular difficulties in terms of robustness for most trackers. While most trackers fail rarely during this attribute, there is observable difference in the accuracy of tracking.

\begin{figure}[h!]
    \centering
    \includegraphics[width=8.5cm]{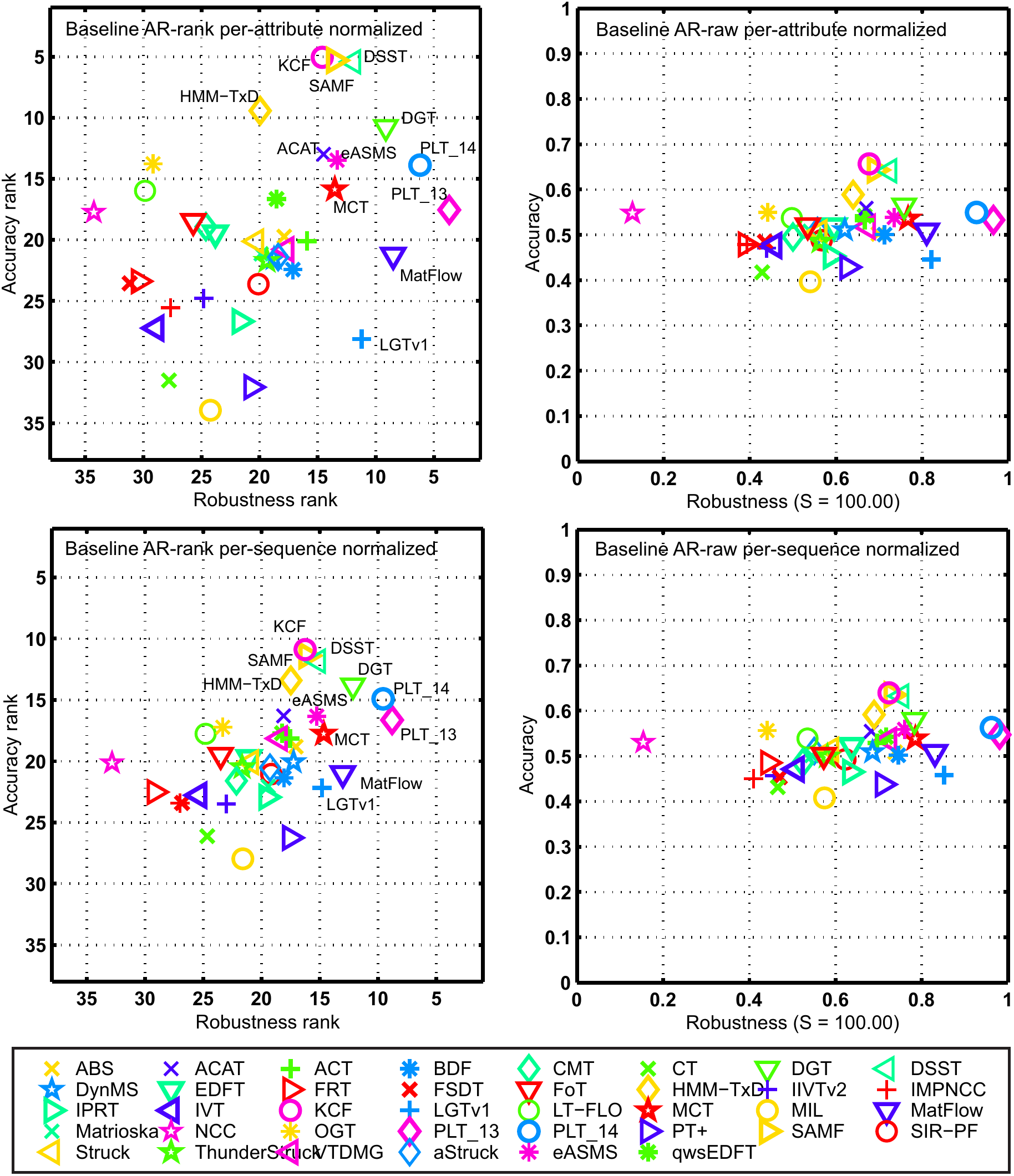}
    \caption{The AR-rank and raw plots for the baseline experiment with per-attribute normalization (upper row) and per-sequence normalization (bottom row).}
    \label{fig:rankingplotsAttr}
\end{figure}

Figure~\ref{fig:rankingplotsAttr} shows the per-visual attribute normalized AR-rank plot for the \textit{baseline} experiment. This plot was obtained by ranking trackers with respect to each attribute and averaging the ranking lists. In Table~\ref{tab:ranking}, these results are denoted as per-attribute normalization. The AR-raw plot in Figure~\ref{fig:rankingplotsAttr} was obtained by averaging per-attribute average raw performance measures. The general layout of the trackers is similar to the sequence-pooled AR plots in Figure~\ref{fig:rankingplots}, but there are differences in local ranks. The reason is that the sequence-pooled plots significantly depend on the distribution of the visual attributes in the dataset. This is confirmed by noting that the most strongly presented attributes in our dataset are camera motion and object motion (Figure~\ref{fig:datasetstats}) and by observing that the structure of the AR-rank plot for the baseline experiment (Figure~\ref{fig:rankingplots}) is very similar to the camera motion and object motion AR-rank plots from Figure~\ref{fig:rankingplotsAttr}. The attribute-normalized AR plots in Figure~\ref{fig:rankingplots2} removes this bias, giving equal importance to all the visual attributes. Averaging the accuracy and robustness ranks in the per-attribute normalization setup, the top performing trackers are DSST, SAMF, KCF, DGT and PLT trackers (see Table~\ref{tab:ranking}). For reference, we also report the results for the sequence-normalized ranking which ranks trackers with respect to each sequence separately and averages the ranking lists. The resulting plots are shown in the bottom row of Figure~\ref{fig:rankingplotsAttr}. Observe that the general distribution of the trackers remains similar to the sequence-pooled plots Figure~\ref{fig:rankingplots}, reflecting the influence of the dominant visual attributes in the dataset. The most apparent difference is that the trackers are less dispersed in the AR-rank space. This is because 25 ranking lists are averaged, indicating that the tracker ranking lists vary over the individual sequences and are consequently pulled to the average rank by averaging.

Note that majority of the tested trackers are highly competitive. This is supported by the fact that the trackers, that are often used as baseline trackers, NCC, MIL, CT,  FRT and IVT, occupy the bottom-left part of the AR-rank plots. Obviously these approaches vary in accuracy and robustness and are thus spread perpendicularly to the bottom-left-to-upper-right diagonal of AR-rank plots. In both experiments, the NCC is the least robust tracker. The Struck, which is often considered a state-of-the-art tracker is positioned in the middle of the AR plots, which further supports the quality of the tested trackers.

\begin{table*}[pt!]\caption{Ranking results of the \textit{baseline} and \textit{bounding box perturbation} experiments without rank normalization (sequence-pooled)  and the \textit{baseline} experiment with per-attribute normalization. The per-accuracy and per-robustness averaged ranks are denoted by A and R, respectively. The top, second and third lowest average ranks are shown in red, blue and green respectively. The last four columns denote tracker properties which are split into: localization (stochastic/deterministic, i.e., S/D); model type (holistic/part-based, i.e., H/P); visual model representation (generative/discriminative, i.e., G/D); scale adaptation (yes/no, i.e., Y/N).}\label{tab:ranking}
\centering\begin{tabular}{l|c|c|c|c|c|c|c|c|c||c|c|c|c}
{Experiment} & \multicolumn{3}{ c| }{ { Baseline}} & \multicolumn{3}{ c| }{ {Region perturbation}} & \multicolumn{3}{ c| }{{Baseline}} & \multicolumn{4}{ c }{{Properties}} \\\hline
{Normalization} & \multicolumn{3}{ c| }{sequence-pooled} & \multicolumn{3}{ c| }{{sequence-pooled}} & \multicolumn{3}{ c| }{per-attribute} & \multicolumn{4}{ c }{ }\\\hline
{Ranking measure} & {A} & {R} & {Avg.} & {A} & {R} & {Avg.} & {A} & {R} & {Avg.} & {Loc.} & {Model} & {Repr.} & {Scale} \\\hline\hline
{DSST~\cite{2014_DANELLJAN_BMVC}} & \second{3.67} & 9.00 & \second{6.33} & \third{5.25} & 9.78 & 7.51 & \third{5.41} & 12.08 & \first{8.75} & D & H & D & Y \\
{SAMF~\cite{2014_LIZHU}} & \first{3.00} & 11.91 & 7.45 & \second{4.00} & 10.70 & 7.35 & \second{5.30} & 13.60 & \second{9.45} & D & P & D & Y \\
{KCF~\cite{Henriques2014}} & \first{3.00} & 12.33 & 7.67 & \second{4.00} & 12.15 & 8.08 & \first{5.05} & 14.67 & \third{9.86} & D & H & D & N \\
{DGT~\cite{2014_CAI}} & \third{4.62} & \third{5.00} & \first{4.81} & \first{3.50} & \second{4.00} & \first{3.75} & 10.76 & 9.13 & 9.95 & D & P & G & Y \\
$\mathrm{PLT}_{14}~\cite{KRISTAN_2014_ECCV}$ & 12.29 & \second{2.00} & \third{7.15} & 10.00 & \first{1.50} & \second{5.75} & 13.88 & \second{6.20} & 10.04 & D & H & D & Y \\
$\mathrm{PLT}_{13}~\cite{Kristan2013a}$ & 17.50 & \first{1.00} & 9.25 & 15.55 & \first{1.50} & 8.52 & 17.54 & \first{3.67} & 10.60 & D & H & D & N \\
{eASMS~\cite{2014_VOJIR}} & 10.00 & 6.80 & 8.40 & 6.00 & \third{6.83} & \third{6.42} & 13.48 & 13.35 & 13.41 & D & H & G & Y \\
{ACAT~\cite{KRISTAN_2014_ECCV}} & 16.00 & 17.54 & 16.77 & 19.54 & 12.15 & 15.85 & 12.99 & 14.58 & 13.79 & D & H & G & Y \\
{HMM-TxD~\cite{KRISTAN_2014_ECCV}} & 5.00 & 16.80 & 10.90 & 6.00 & 12.45 & 9.23 & 9.43 & 19.96 & 14.70 & D & P & G & Y \\
{MCT~\cite{2014_DUFFNER}} & 17.50 & 7.83 & 12.67 & 20.07 & 11.70 & 15.89 & 15.88 & 13.61 & 14.74 & S & H & G & Y \\
{MatFlow~\cite{KRISTAN_2014_ECCV}} & 21.54 & \third{5.00} & 13.27 & 19.54 & 12.15 & 15.85 & 21.25 & \third{8.52} & 14.88 & D & P & G & N \\
{qwsEDFT~\cite{2014_FELSBERG}} & 17.50 & 16.92 & 17.21 & 19.00 & 21.75 & 20.38 & 16.65 & 18.50 & 17.58 & D & H & G & N \\
{ACT~\cite{2014_DANELLJAN}} & 19.42 & 14.62 & 17.02 & 22.00 & 12.42 & 17.21 & 20.08 & 15.92 & 18.00 & D & H & D & N \\
{ABS~\cite{KRISTAN_2014_ECCV}} & 17.50 & 16.92 & 17.21 & 13.45 & 12.15 & 12.80 & 19.72 & 17.93 & 18.83 & D & H & G & Y \\
{VTDMG~\cite{KRISTAN_2014_ECCV}} & 17.50 & 15.69 & 16.60 & 19.00 & 11.00 & 15.00 & 20.77 & 17.69 & 19.23 & D & H & G & N \\
{LGT~\cite{Cehovin2013}} & 28.63 & 5.75 & 17.19 & 23.81 & \second{4.00} & 13.91 & 28.12 & 11.28 & 19.70 & S & P & G & Y \\
{BDF~\cite{2014_MARESCA_a}} & 23.50 & 15.69 & 19.60 & 22.29 & 15.45 & 18.87 & 22.42 & 17.10 & 19.76 & D & P & G & N \\
{aStruck~\cite{KRISTAN_2014_ECCV}} & 22.50 & 20.45 & 21.48 & 22.29 & 25.64 & 23.96 & 21.41 & 18.43 & 19.92 & D & P & D & N \\
{DynMS~\cite{KRISTAN_2014_ECCV}} & 18.54 & 15.69 & 17.12 & 19.54 & 15.58 & 17.56 & 21.54 & 18.80 & 20.17 & S & H & G & Y \\
{Struck~\cite{hare2011iccv}} & 19.58 & 24.60 & 22.09 & 22.00 & 20.44 & 21.22 & 20.11 & 20.30 & 20.21 & D & H & D & N \\
{Matrioska~\cite{Maresca2013}} & 21.54 & 18.33 & 19.94 & 21.50 & 27.62 & 24.56 & 21.15 & 19.92 & 20.53 & D & P & G & N \\
{TStruck~\cite{hare2011iccv}} & 21.54 & 25.64 & 23.59 & 22.00 & 20.44 & 21.22 & 21.71 & 19.38 & 20.55 & D & H & D & N \\
{OGT~\cite{2014_NAM}} & 12.06 & 29.78 & 20.92 & 16.50 & 30.58 & 23.54 & 13.76 & 29.13 & 21.44 & S & H & G & N \\
{EDFT~\cite{Felsberg2013vot}} & 18.54 & 24.43 & 21.49 & 21.50 & 24.70 & 23.10 & 19.43 & 23.71 & 21.57 & D & H & G & N \\
{CMT~\cite{Nebehay2014WACV}} & 20.17 & 27.44 & 23.81 & 24.72 & 27.30 & 26.01 & 18.93 & 24.53 & 21.73 & D & P & G & Y \\
{SIR-PF~\cite{KRISTAN_2014_ECCV}} & 23.50 & 18.50 & 21.00 & 20.07 & 24.70 & 22.39 & 23.62 & 20.13 & 21.88 & S & H & G & N \\
{FoT~\cite{vojir2011cvww}} & 21.00 & 27.44 & 24.22 & 23.32 & 31.20 & 27.26 & 18.48 & 25.67 & 22.07 & D & P & G & Y \\
{LT-FLO~\cite{Lebeda2013vot}} & 17.50 & 30.50 & 24.00 & 20.07 & 31.20 & 25.64 & 15.98 & 29.85 & 22.91 & S & P & G & Y \\
{IPRT~\cite{KRISTAN_2014_ECCV}} & 26.67 & 22.33 & 24.50 & 23.81 & 23.78 & 23.80 & 26.68 & 21.72 & 24.20 & S & H & G & N \\
{IIVTv2~\cite{KRISTAN_2014_ECCV}} & 29.35 & 30.67 & 30.01 & 26.18 & 28.17 & 27.17 & 24.79 & 24.81 & 24.80 & D & P & G & Y \\
{NCC~\cite{2001_BRIECHLE}} & 17.50 & 38.00 & 27.75 & 22.29 & 38.00 & 30.14 & 17.74 & 34.25 & 26.00 & D & H & G & N \\
{PT+~\cite{KRISTAN_2014_ECCV}} & 32.64 & 15.69 & 24.16 & 27.84 & 13.67 & 20.75 & 32.05 & 20.68 & 26.36 & D & P & G & Y \\
{IMPNCC~\cite{KRISTAN_2014_ECCV}} & 29.73 & 33.25 & 31.49 & 32.42 & 31.71 & 32.07 & 25.56 & 27.68 & 26.62 & D & H & G & Y \\
{FRT~\cite{Adam2006}} & 21.00 & 35.00 & 28.00 & 23.81 & 36.00 & 29.91 & 23.38 & 30.39 & 26.89 & D & P & G & N \\
{FSDT~\cite{KRISTAN_2014_ECCV}} & 31.50 & 33.40 & 32.45 & 23.32 & 30.73 & 27.02 & 23.55 & 31.16 & 27.36 & D & H & D & Y \\
{IVT~\cite{Ross2008}} & 28.05 & 33.14 & 30.60 & 28.35 & 31.20 & 29.77 & 27.23 & 28.90 & 28.06 & D & H & G & Y \\
{MIL~\cite{babenko2011tpami}} & 34.25 & 28.38 & 31.31 & 35.75 & 30.10 & 32.92 & 33.95 & 24.20 & 29.08 & D & H & D & N \\
{CT~\cite{zhang2012eccv}} & 32.64 & 33.14 & 32.89 & 29.00 & 30.88 & 29.94 & 31.51 & 27.79 & 29.65 & D & H & D & N \\
\end{tabular}
\end{table*}


Next, we have ranked the individual types of visual degradation according to the tracking difficulty they present to the tested trackers. The expected number of failures per hundred frames was computed on each attribute for all trackers. The median of these per visual attribute was taken as a measure of tracking difficulty (see Table~\ref{tab:properties}). The properties that present most difficulty are occlusion, motion change and size change, followed by camera motion and illumination change. Subsequences that do not contain any specific attribute (neutral) present little difficulty for the trackers in general as most trackers do not fail on such intervals.

\begin{table}[h!]\caption{Tracking difficulty for the six visual attributes: camera motion (CM), illumination change (IC), occlusion (OC), object size change (SC), object motion change (MC) and neutral (NE).}\label{tab:properties}
\centering{\small
\begin{tabular}{ c|c c c c c c }
&\textbf{CM}&\textbf{IC}&\textbf{OC}&\textbf{SC}&\textbf{MC}& \textbf{NE}  \\\hline
\textbf{Exp. failures} & 0.55 & 0.42 & \first{1.13} & \third{0.74} & \second{0.79} & 0.00 \\
\end{tabular}
}
\end{table}

\subsection{Results of Sequence analysis}

\begin{figure}
\includegraphics[width=8.5cm]{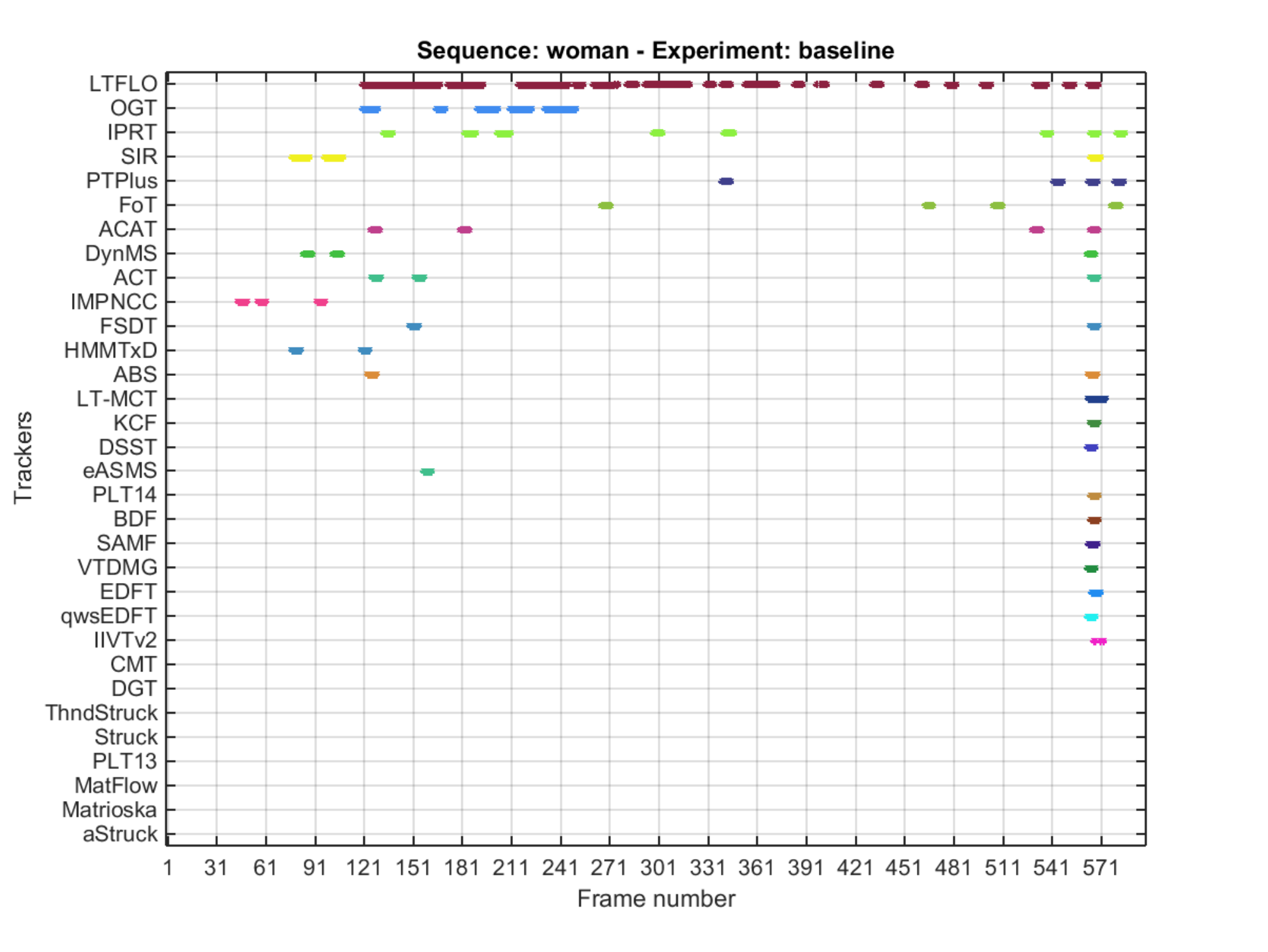}
\caption{The scatter plot for the {\em woman} sequence shows the failures for each tracker w.r.t. frame number.}
\label{fig:scatterplot}
\end{figure}

\begin{figure}
\includegraphics[width=8.5cm]{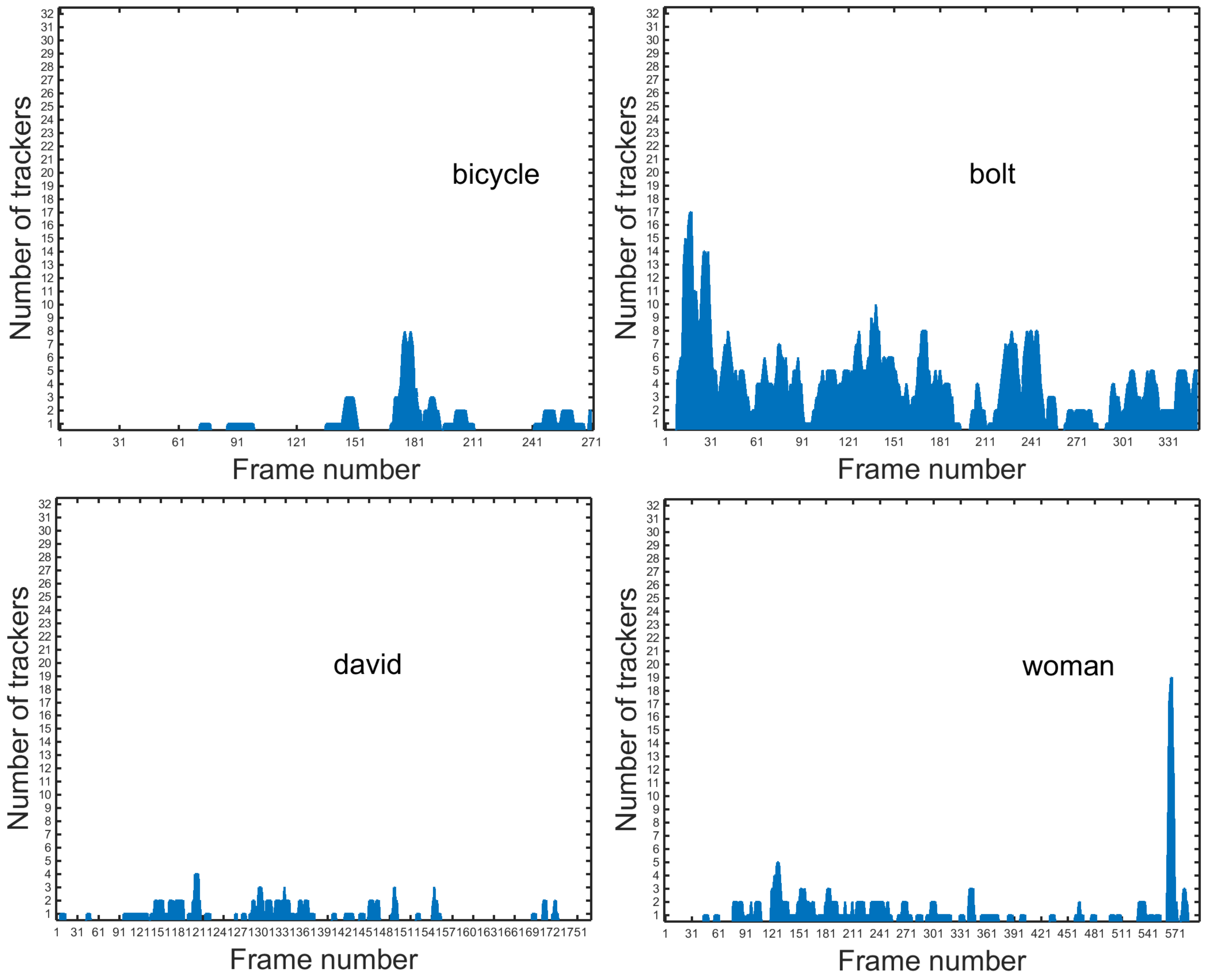}
\caption{\label{fig:seqAnalysis} Difficulty curves for the {\em bicycle}, {\em bolt}, {\em david}, and {\em woman} sequences.}
\end{figure}

A further analysis was conducted to gain an insight into the dataset from a tracker perspective. For each sequence we have analyzed if a particular tracker failed at least once at a particular frame~(Figure~\ref{fig:scatterplot}). By counting how many trackers failed at each frame, the level of difficulty can be visualized by the difficulty curve for each sequence~(Figure~\ref{fig:seqAnalysis}). From these curves two measures of sequence difficulty are derived:~\textit{area} and \textit{max}. The \textit{area} is a sum of frame-wise values from the difficulty curve normalized by the number of frames, while the \textit{max} is the maximum on this curve. The former indicates the average level of difficulty of a sequence, and the latter reflects the difficulty of the most difficult part in the sequence. Table~\ref{seq-measures} summarizes the \textit{area} and \textit{max} values for all sequences. A high value of the area suggests that such sequence is challenging in a considerable number of frames. For example, the \textit{area} for the {\em david} sequence is smaller than the \textit{area} for the {\em woman} sequence, which suggests that {\em david} sequence is less challenging that the {\em woman} sequence. A large {\em max} indicates the presence of difficult frames.
For example, a significant peak in the {\em woman} sequence~(frame~566) suggests that this sequence contains a subsequence around this frame which is challenging to most of the trackers. In case of drunk sequence, the corresponding max value is~$3$~(see Table~\ref{seq-measures}), thus almost all trackers successfully track the target.

\begin{table}
{\footnotesize
  \begin{center}
    \begin{tabular}{l|cccc}
      Sequence    			& area 	& max 	& frame & difficulty		\\
     \hline
      {\em motocross}		& 5.92      & 19 				&  39					&	hard						\\
      {\em hand2}				& 5.65      & 24 				& 167					&	hard						\\
      {\em diving}			&	4.85			&	15				&	195					&	hard					\\
      {\em fish2}				&	4.59			&	16				&	 35					&	hard					\\
      {\em bolt}				&	4.14			&	17				&	 17					&	hard					\\
      {\em hand1}				&	3.23			&	15				&	 51					&	hard		\\
      {\em fish1}				&	2.94			&	16				&	 39					&	interm		\\
      {\em fernando}		&	2.78			&	19				&	292					&	interm		\\
      {\em gymnastics}	&	2.59			&	19				&	 97					&	interm		\\
      {\em torus}				&	2.26			&	 9				&	146					&	interm		\\
      {\em skating}			&	2.12			&	 9				&	312					&	interm		\\
      {\em trellis}			&	1.58			&	10				&	391					&	interm./easy						\\
      {\em basketball}	&	1.43			&	11				&	668					&	interm./easy						\\
      {\em tunnel}			&	1.27			&	 6				&	493					&	interm./easy					\\
      {\em sunshade}		&	1.24			&	12				&	114					&	interm./easy						\\
      {\em jogging}			&	1.12			&	28				&	 77					&	interm./easy						\\
      {\em woman}				&	1.05			&	19				&	566					&	interm./easy						\\
      {\em bicycle}			&	0.75			&	 8				&	176					&	easy						\\
      {\em david}				&	0.60			&	 4				&	200					&	easy						\\
      {\em ball}				&	0.47			&	 7				&	189					&	easy						\\
      {\em sphere}			&	0.41			&	 3				&	 33					&	easy						\\
      {\em car}					&	0.25			&	 7				&	170					&	easy						\\
      {\em drunk}				&	0.11			&	 3				&	248					&	easy						\\
      {\em surfing}			&	0.04			&	 1				&	178					&	easy						\\
      {\em polarbear}		&	0.00			&	 0				&	 1 					&	easy						\\
    \end{tabular}
  \end{center}
}
\caption{Sequence difficulty from tracking perspective. The table shows for each sequence the average number of per-frame failed trackers (area), the frame (frame) where maximum number (max) of trackers simultaneously failed and the difficulty level (difficulty).}
  \label{seq-measures}
\end{table}
 
Using the \textit{area} measure the sequences were labeled by the following four levels of difficulty: Hard~(area greater than~$3.00$), intermediate~(area between~$3.00$ and~$2.00$), intermediate/easy~(area between~$1.00$ and~$1.00$) and easy~(area less than~$1.00$)~(see Table~\ref{seq-measures}). These levels were defined by manually clustering the areas into four clear clusters. Surprisingly, the {\em david} sequence (Figure~\ref{fig:seqAnalysis}) shows a small \textit{area} in this study, although the sequence is usually considered in the community to be challenging and it is commonly referred in the literature. One explanation might be that the trackers are over-fitted to this sequence since it is so often used in evaluation and development. An alternative explanation might be that the sequence is actually not very challenging for tracking, but appears to be to a human observer. The popularity would then be explained by the fact that it is appealing to demonstrate good tracking performance on a sequence that appears difficult, even though it might not be. The analysis also shows that the {\em motocross}, {\em hand2}, {\em diving}, {\em fish2}, {\em bolt} and {\em hand1} are the most challenging sequences. Most of the difficulties in these sequences arise from changes in camera and object motion as well as from rapid changes in object size. 
For example, {\em motocross} is hard because all three aforementioned nuisances occur simultaneously while the {\em hand2} sequence shows challenging pose variations of the person's hand. The {\em diving} sequence shows significant changes in object size, in {\em bolt} sequence both motions camera and object occur simultaneously, while the {\em fish2} sequence shows challenging pose variations of the object.
 
Easy to intermediate sequences might remain valuable for tracker comparison as these sequences still conceal challenges in particular frames. These sequences are identified by considering \textit{max} in Table~\ref{seq-measures}. 
For example, almost all trackers fail at frame~$77$ of the {\em jogging} sequence. A closer look at this frame and previous frames shows a complete occlusion of the object. Similarly, the {\em woman} sequence at frame~$566$ (Figure~\ref{fig:seqAnalysis}) contains camera zooming which makes $19$ out of $38$~trackers fail. The {\em bicycle} sequence also shows a peak in the difficulty curve at frame~$176$~(Figure~\ref{fig:seqAnalysis}). In this part of the sequence, an object is occluded, which is immediately followed by a shadow cast over the target. A significant peak is also present in the \textit{bolt} sequence (Figure~\ref{fig:seqAnalysis}) at frame~$17$, at which many trackers fail. A closer look at the frame and its neighbouring frames shows a significant object motion between the frames as a cause of failures.

\section{Conclusion}\label{sec:Conclusion}

In this paper a novel tracker performance evaluation methodology was presented. Requirements for the performance measures, the dataset and the evaluation system are defined and a new evaluation methodology is proposed which aims at a simple, easily interpretable, tracker comparison. The proposed methodology is the first of its kind to account for the tracker equivalence by considering statistical significance and practical differences. A new dataset and a cross-platform-compatible evaluation system were presented. The dataset consists of 25 color sequences, which are per-frame annotated by visual attributes and rotated boxes. Effects of re-initialization and per-frame annotation are studied theoretically and the theoretical predictions are verified with experiments. The novel performance evaluation was applied to comparison of 38 trackers, making it the largest benchmark to date. Using the benchmark, the dataset was analyzed from perspective of per-sequence and per-visual-attribute tracking difficulty. The raw results of all trackers are publicly available from the VOT homepage for reproduction of the results in this paper and to allow comparison with new trackers.

The results of an exhaustive analysis show that trackers tend to specialize either for robustness or accuracy. None of the trackers consistently outperformed the others by all measures at all sequence attributes. The top-performing trackers include trackers with holistic as well as part-based visual models. There is some evidence that robustness is achieved by discriminative learning where variants of structured SVM, e.g. PLT, seem promising. Variants of segmentation appear to play a beneficial role in tracking with noisy initializations. This is evident in favorable performance of trackers DGT and PLTs in the noise experiment. But relying strongly on segmentation  reduces performance when color significantly changes which is seen in significant deterioration of the DGT on illumination change. Estimation of few parameters likely increases tracking robustness at reduced accuracy. Attribute-wise analysis shows that motion prediction significantly improves performance during dynamic target motion. Results show that evaluating trackers by pooling results from sequences largely depends on the types of attributes that dominate the dataset. A per-visual-attribute analysis and attribute normalization in final ranking is thus beneficial to remove this bias. Most of the tested trackers outperform standard baselines and perform favorably to common state-of-the-art such as Struck, making the benchmark quite challenging.
 
The per-attribute analysis of the new dataset showed that the visual attributes that are most challenging to trackers are occlusion, motion change and size change. Sequence-wise analysis showed that some sequences are challenging on average, other sequences are very challenging at particular frames, and some of them are well tackled by all the trackers. An interesting find is that one particular sequence (David), which is usually assumed challenging in the tracking community, seems not to be according to the presented analysis, as trackers rarely fail on this sequence.

Establishing standard datasets and evaluation methodology tends to result in significant short-term advances in the field, but it can also have negative effects, leading to empoverished specter of approaches that get put forward in the long run~\cite{Torralba_CVPR2011}. Evaluation is often reduced to a single performance score, which might lead to degradation in research. The primary goal of the authors, i.e., coming up with new tracking concepts, shifts to increasing a single performance score, and this is further enforced by pre-occupied reviewers that may find appealing to base their decision on this single score as well. We would like to explicitly warn against this. In practical experiments we are in fact comparing performance of various implementations rather than concepts. Implementations sometimes contain tweaks that improve performance, while often being left out from the original papers in interest of purity of the theory. 

We also point out that the notion of a ''best'' tracker varies with the tracker application. For example, sports analytics applications, which sports scientists use for player accelerations and velocity analysis, crucially depend on the quality of the estimated player position and do not require autonomous real-time performance. Thus user intervention for tracker reinitialization is allowed at any point. In such applications a highly accurate tracker is required, but robustness is only desired, i.e., an accurate non-robust tracker would be preferred over a robust but inaccurate tracker. But other applications in which tracking autonomy is critical, a robust tracker would be preferred over an accurate but non-robust tracker. The presented methodology allows identifying these characteristics and their variation w.r.t. the visual attributes which goes beyond the related methodologies.

We believe that it is difficult to overfit a tracker to a visually diverse dataset, but tuning parameters may very likely contribute to higher ranks. Related works like~\cite{Li2015tpami} suggest splitting the dataset into training and testing sequences, making all sequences available, but only providing the annotations for the training sequences. The evaluation is then performed by running the tracker on the test set and uploading the results to an online service that checks the results against the unpublished ground truth. One problem with such an approach is that re-initialization at failure becomes impossible, since the test-data ground truth is censored, thus reducing the strength of the performance measures. But a conceptual
problem lies in the assumption that the ''unpublished'' ground truth cannot be re-produced. In fact, if the annotation rules are followed faithfully, the researchers can easily annotate the ground truth in the censored part of the dataset and this annotation will be equally valid as the unpublished. So if overfitting would be possible, censoring the ground truth would introduce even a larger bias in the results in favor of researchers that simply spend time re-annotating the test dataset.
 
Because of the unavoidable dependence on implementation and efforts spent in adjusting the tracker parameters, care has to be taken when deciding for or against a new tracker based on performance scores. One approach might be to apply a comparative evaluation to position a new tracking approach against a set of standard baseline implementations using a single ranking experiment, use detailed analysis with respect to different visual attributes and put further focus on the theory.

Our future work will focus on revising and carefully enriching the dataset, continually improving the tracker evaluation methodology and, through further organization of the VOT challenges, pushing towards a standardised tracker comparison.

%
%
%
\appendices
\section{Derivation of NOR and WIR statistics}\label{app:DeriveNORWIR}

The derivation of the results in the bias-variance analysis of the NOR and WIR overlap estimators in equations (\ref{eq:otb_est_overall}-\ref{eq:vot_est_overall_last}) from Section~\ref{sec:theoreticalEval} is outlined here. 
Recall that the tracking accuracy is measured by $M = {1 \over N} \sum\nolimits_{j=1:N} M_{j}$ where $M_{j}={1 \over N_\mathrm{s}}\sum\nolimits_{i=1:N_\mathrm{s}} o_{ij}$ is the tracking accuracy at $j$-th sequence. This accuracy is a random variable governed by a mixture model $M_{j} \sim p_f(\mu_f, \sigma_f^2) p + (1-p) p_s(\mu_s, \sigma_s^2)$ where $p_f(\cdot)$ and $p_s(\cdot)$ are distributions with mean and variance ($\mu$, $\sigma$) describing the statistics of the average overlap in case a failure in the sequence occurs or not, respectively. The mean and variance of the mixture model are
\begin{eqnarray} 
	\langle M_j \rangle = p \mu_f + (1-p) \mu_s \label{eq:gen_mu}  \\
 	\mathrm{var}(M_j) = p\sigma_f^2 + (1-p)\sigma_s^2 + p(1-p)(\mu_f - \mu_s)^2. \label{eq:gen_var}
\end{eqnarray}

We will first consider the NOR scenario. In case a failure does not occur, the parameters of $p_s(\mu_{\mathrm{NOR}s}, \sigma_{\mathrm{NOR}s}^2)$ are trivially computed, i.e.,
\begin{eqnarray}  
	\mu_{\mathrm{NOR}s} = \mu_A {~;~} \sigma_{\mathrm{NOR}s}^2 = {1 \over N_\mathrm{s}} \sigma_A^2. 
\end{eqnarray}
In case of failure, the overlap drops to zero after $N_\mathrm{s}\alpha_j$ frames, thus the mean value of $p_f(\mu_{\mathrm{NOR}f}, \sigma_{\mathrm{NOR}f}^2)$ is
\begin{equation}
	\mu_{\mathrm{NOR}f} = \langle \alpha_j \mu_A \rangle = {1 \over 2} \mu_A.
\end{equation}
The variance $\sigma_{\mathrm{NOR}f}^2$ is computed by application of the total variance law, yielding
\begin{equation}
	\sigma_{\mathrm{NOR}f}^2 = {1 \over 2 N_\mathrm{s}} \sigma_A^2 + {1 \over 12} \mu_A^2.
\end{equation}
Plugging these results into (\ref{eq:gen_mu},\ref{eq:gen_var}) yields equations (5) and (6) in the paper.

In the WIR scenario, the tracker is reset after failure and $\Delta$ frames after the reset are ignored in computation of the accuracy. It is easy to show the following equivalence
\begin{eqnarray} 
	\mu_{\mathrm{WIR}s} = \mu_A {~;~}
    \sigma_{\mathrm{WIR}s}^2 =  {1 \over N_\mathrm{s}} \sigma_A^2\\
    \mu_{\mathrm{WIR}f} = \mu_A {~;~}
    \sigma_{\mathrm{WIR}f}^2 =  {1 \over (N_\mathrm{s}-\Delta)} \sigma_A^2.
\end{eqnarray}
Plugging these into (\ref{eq:gen_mu},\ref{eq:gen_var}) yields equations (7) and (8) in the paper.

%

\ifCLASSOPTIONcompsoc
  \section*{Acknowledgments}
\else
  \section*{Acknowledgment}
\fi

This work was supported in part by the following research programs and projects: Slovenian research agency research programs and projects \mbox{P2-0095}, \mbox{P2-0214}, J2-4284, J2-3607, the EU project EPiCS~(grant agreement no 257906), the CTU Project SGS15/155/OHK3/2T/13 and by The Czech Science Foundation Project GACR P103/12/G084.

\ifCLASSOPTIONcaptionsoff
  \newpage
\fi

\bibliographystyle{IEEEtran}
\bibliography{bib/tpami_bibliography}

\begin{thebibliography}{10}
\providecommand{\url}[1]{#1}
\csname url@samestyle\endcsname
\providecommand{\newblock}{\relax}
\providecommand{\bibinfo}[2]{#2}
\providecommand{\BIBentrySTDinterwordspacing}{\spaceskip=0pt\relax}
\providecommand{\BIBentryALTinterwordstretchfactor}{4}
\providecommand{\BIBentryALTinterwordspacing}{\spaceskip=\fontdimen2\font plus
\BIBentryALTinterwordstretchfactor\fontdimen3\font minus
  \fontdimen4\font\relax}
\providecommand{\BIBforeignlanguage}[2]{{%
\expandafter\ifx\csname l@#1\endcsname\relax
\typeout{** WARNING: IEEEtran.bst: No hyphenation pattern has been}%
\typeout{** loaded for the language `#1'. Using the pattern for}%
\typeout{** the default language instead.}%
\else
\language=\csname l@#1\endcsname
\fi
#2}}
\providecommand{\BIBdecl}{\relax}
\BIBdecl

\bibitem{Gavrila99}
D.~M. Gavrila, ``The visual analysis of human movement: A survey,'' \emph{Comp.
  Vis. Image Understanding}, vol.~73, no.~1, pp. 82--98, 1999.

\bibitem{Moeslund2001}
T.~B. Moeslund and E.~Granum, ``A survey of computer vision-based human motion
  capture,'' \emph{Comp. Vis. Image Understanding}, vol.~81, no.~3, pp.
  231--268, March 2001.

\bibitem{Gabriel03}
P.~Gabriel, J.~Verly, J.~Piater, and A.~Genon, ``The state of the art in
  multiple object tracking under occlusion in video sequences,'' in \emph{Proc.
  Advanced Concepts for Intelligent Vision Systems}, 2003, pp. 166--173.

\bibitem{Hu2004}
W.~Hu, T.~Tan, L.~Wang, and S.~Maybank, ``A survey on visual surveillance of
  object motion and behaviors,'' \emph{IEEE Trans. Systems, Man and
  Cybernetics, C}, vol.~34, no.~30, pp. 334--352, 2004.

\bibitem{Moeslund2006}
T.~B. Moeslund, A.~Hilton, and V.~Kruger, ``A survey of advances in
  vision-based human motion capture and analysis,'' \emph{Comp. Vis. Image
  Understanding}, vol. 103, no. 2-3, pp. 90--126, November 2006.

\bibitem{Yilmaz2006}
A.~Yilmaz and M.~Shah, ``Object tracking: A survey,'' \emph{Journal ACM
  Computing Surveys}, vol.~38, no.~4, 2006.

\bibitem{Li2013}
X.~Li, W.~Hu, C.~Shen, Z.~Zhang, A.~R. Dick, and A.~Van~den Hengel, ``A survey
  of appearance models in visual object tracking,'' \emph{arXiv:1303.4803
  [cs.CV]}, 2013.

\bibitem{Young2005}
D.~P. Young and J.~M. Ferryman, ``{PETS M}etrics: On-line performance
  evaluation service,'' in \emph{ICCCN '05 Proceedings of the 14th
  International Conference on Computer Communications and Networks}, 2005, pp.
  317--324.

\bibitem{GoyetteCVPR12}
N.~Goyette, P.~M. Jodoin, F.~Porikli, J.~Konrad, and P.~Ishwar,
  ``Changedetection.net: A new change detection benchmark dataset.'' in
  \emph{CVPR Workshops}.\hskip 1em plus 0.5em minus 0.4em\relax IEEE, 2012, pp.
  1--8.

\bibitem{Phillips2000}
P.~J. Phillips, H.~Moon, S.~A. Rizvi, and P.~J. Rauss, ``The feret evaluation
  methodology for face-recognition algorithms,'' \emph{IEEE Trans. Pattern
  Anal. Mach. Intell.}, vol.~22, no.~10, pp. 1090--1104, 2000.

\bibitem{Kasturi_TPAMI_2009}
R.~Kasturi, D.~B. Goldgof, P.~Soundararajan, V.~Manohar, J.~S. Garofolo,
  R.~Bowers, M.~Boonstra, V.~N. Korzhova, and J.~Zhang, ``Framework for
  performance evaluation of face, text, and vehicle detection and tracking in
  video: Data, metrics, and protocol,'' \emph{IEEE Trans. Pattern Anal. Mach.
  Intell.}, vol.~31, no.~2, pp. 319--336, 2009.

\bibitem{Geiger2012CVPR}
A.~Geiger, P.~Lenz, and R.~Urtasun, ``Are we ready for autonomous driving? the
  kitti vision benchmark suite,'' in \emph{CVPR}.\hskip 1em plus 0.5em minus
  0.4em\relax IEEE, 2012, pp. 3354--3361.

\bibitem{Bernardin2008}
K.~Bernardin and R.~Stiefelhagen, ``Evaluating multiple object tracking
  performance: The clear mot metrics,'' \emph{EURASIP Journal on Image and
  Video Processing, Special Issue on Video Tracking in Complex Scenes for
  Surveillance Applications}, vol. 2008, 2008.

\bibitem{Karasulu2011}
B.~Karasulu and S.~Korukoglu, ``A software for performance evaluation and
  comparison of people detection and tracking methods in video processing,''
  \emph{Multimedia Tools and Applications}, vol.~55, no.~3, pp. 677--723, 2011.

\bibitem{Salti2012}
S.~Salti, A.~Cavallaro, and L.~Di~Stefano, ``Adaptive appearance modeling for
  video tracking: Survey and evaluation,'' \emph{IEEE Trans. Image Proc.},
  vol.~21, no.~10, pp. 4334 -- 4348, 2012.

\bibitem{Wu2013}
Y.~Wu, J.~Lim, and M.~H. Yang, ``Online object tracking: A benchmark,'' in
  \emph{Comp. Vis. Patt. Recognition}, 2013.

\bibitem{Smeulders2013}
A.~W.~M. Smeulders, D.~M. Chu, R.~Cucchiara, S.~Calderara, A.~Dehghan, and
  M.~Shah, ``{Visual Tracking: an Experimental Survey},'' \emph{IEEE Trans.
  Pattern Anal. Mach. Intell.}, 2013.

\bibitem{Pang2013}
Y.~Pang and H.~Ling, ``Finding the best from the second bests -- inhibiting
  subjective bias in evaluation of visual tracking algorithms,'' in \emph{Int.
  Conf. Computer Vision}, 2013.

\bibitem{Kristan2013}
M.~Kristan and L.~\v{C}ehovin, \emph{Visual Object Tracking Challenge (VOT2013)
  Evaluation Kit}, Visual Object Tracking Challenge, 2013.

\bibitem{Kristan2014}
M.~Kristan, R.~Pflugfelder, A.~Leonardis, J.~Matas, F.~Porikli, L.~Cehovin,
  G.~Nebehay, G.~Fernandez, and T.~Vojir, ``The vot2013 challenge: overview and
  additional results,'' in \emph{Computer Vision Winter Workshop}, 2014.

\bibitem{MOTChallenge2015}
\BIBentryALTinterwordspacing
L.~Leal-Taix\'{e}, A.~Milan, I.~Reid, S.~Roth, and K.~Schindler,
  ``{MOTC}hallenge 2015: {T}owards a benchmark for multi-target tracking,''
  \emph{arXiv:1504.01942 [cs]}, Apr. 2015, arXiv: 1504.01942. [Online].
  Available: \url{http://arxiv.org/abs/1504.01942}
\BIBentrySTDinterwordspacing

\bibitem{Fleuret08a}
F.~Fleuret, J.~Berclaz, R.~Lengagne, and P.~Fua, ``{Multi-Camera People
  Tracking with a Probabilistic Occupancy Map},'' \emph{IEEE Transactions on
  Pattern Analysis and Machine Intelligence}, vol.~30, no.~2, pp. 267--282,
  February 2008.

\bibitem{Kristan2013a}
M.~Kristan, R.~Pflugfelder, A.~Leonardis, J.~Matas, F.~Porikli, L.~\v{C}ehovin,
  G.~Nebehay, G.~Fernandez, T.~Vojir, A.~Gatt, A.~Khajenezhad, A.~Salahledin,
  A.~Soltani-Farani, A.~Zarezade, A.~Petrosino, A.~Milton, B.~Bozorgtabar,
  B.~Li, C.~S. Chan, C.~Heng, D.~Ward, D.~Kearney, D.~Monekosso, H.~C.
  Karaimer, H.~R. Rabiee, J.~Zhu, J.~Gao, J.~Xiao, J.~Zhang, J.~Xing, K.~Huang,
  K.~Lebeda, L.~Cao, M.~E. Maresca, M.~K. Lim, M.~E. Helw, M.~Felsberg,
  P.~Remagnino, R.~Bowden, R.~Goecke, R.~Stolkin, S.~Y. Lim, S.~Maher,
  S.~Poullot, S.~Wong, S.~Satoh, W.~Chen, W.~Hu, X.~Zhang, Y.~Li, and Z.~Niu,
  ``{The Visual Object Tracking VOT2013 challenge results},'' in \emph{ICCV
  Workshops}, 2013, pp. 98--111.

\bibitem{KRISTAN_2014_ECCV}
M.~Kristan, R.~Pflugfelder, A.~Leonardis, J.~Matas, L.~\v{C}ehovin, G.~Nebehay,
  T.~Vojir, G.~Fernandez, A.~Luke\v{z}i\v{c}, A.~Dimitriev, A.~Petrosino,
  A.~Saffari, B.~Li, B.~Han, C.~Heng, C.~Garcia, D.~Panger\v{s}i\v{c},
  G.~H\"{a}ger, F.~Shahbaz~Khan, F.~Oven, H.~Possegger, H.~Bischof, H.~Nam,
  J.~Zhu, J.~Li, J.~Y. Choi, J.-W. Choi, J.~a. Henriques, J.~{V}an~de {W}eijer,
  J.~Batista, K.~Lebeda, K.~\"{O}fj\"{a}ll, K.~M. Yi, L.~Qin, L.~Wen, M.~E.
  Maresca, M.~Danelljan, M.~Felsberg, M.-M. Cheng, P.~Torr, Q.~Huang,
  R.~Bowden, S.~Hare, S.~YueYing~Lim, S.~Hong, S.~Liao, S.~Hadfield, S.~Li,
  S.~Duffner, S.~Golodetz, T.~Mauthner, V.~Vineet, W.~Lin, Y.~Li, Y.~Qi,
  Z.~Lei, and Z.~Niu, ``{The Visual Object Tracking VOT2014 challenge
  results},'' in \emph{ECCV2014 Workshops, Workshop on visual object tracking
  challenge}, 2014, pp. 98--111.

\bibitem{Ross2008}
D.~A. Ross, J.~Lim, R.~S. Lin, and M.~H. Yang, ``Incremental learning for
  robust visual tracking.'' \emph{Int. J. Comput. Vision}, vol.~77, no. 1-3,
  pp. 125--141, 2008.

\bibitem{Li2011}
H.~Li, C.~Shen, and Q.~Shi, ``Real-time visual tracking using compressive
  sensing.'' in \emph{Comp. Vis. Patt. Recognition}.\hskip 1em plus 0.5em minus
  0.4em\relax IEEE, 2011, pp. 1305--1312.

\bibitem{Kwon2009}
J.~Kwon and K.~M. Lee, ``Tracking of a non-rigid object via patch-based dynamic
  appearance modeling and adaptive basin hopping monte carlo sampling.'' in
  \emph{Comp. Vis. Patt. Recognition}.\hskip 1em plus 0.5em minus 0.4em\relax
  IEEE, 2009, pp. 1208--1215.

\bibitem{Kristan2008b}
M.~Kristan, J.~Per{\v s}, M.~Per{\v s}e, and S.~Kova{\v c}i{\v c},
  ``Closed-world tracking of multiple interacting targets for indoor-sports
  applications,'' \emph{Comput. Vision Image Understanding}, vol. 113, no.~5,
  pp. 598--611, May 2009.

\bibitem{Kristan2010b}
M.~Kristan, S.~Kovacic, A.~Leonardis, and J.~Per{\v s}, ``A two-stage dynamic
  model for visual tracking.'' \emph{IEEE Transactions on Systems, Man, and
  Cybernetics, Part B}, vol.~40, no.~6, pp. 1505--1520, 2010.

\bibitem{Nawaz2013}
T.~Nawaz and A.~Cavallaro, ``A protocol for evaluating video trackers under
  real-world conditions.'' \emph{IEEE Trans. Image Proc.}, vol.~22, no.~4, pp.
  1354--1361, 2013.

\bibitem{Carvalho2012}
P.~Carvalho, J.~S. Cardoso, and L.~Corte-Real, ``Filling the gap in quality
  assessment of video object tracking,'' \emph{Image and Vision Computing},
  vol.~30, no.~9, pp. 630--640, 2012.

\bibitem{Cehovin2014wacv}
L.~\v{C}ehovin, M.~Kristan, and A.~Leonardis, ``Is my new tracker really better
  than yours?'' in \emph{IEEE WACV2014}, 2014.

\bibitem{CehovinTracMeas2015}
L.~{\v C}ehovin, A.~Leonardis, and M.~Kristan, ``Visual object tracking
  performance measures revisited,'' \emph{arXiv:1502.05803 [cs.CV]}, 2015.

\bibitem{Everingham2014}
M.~Everingham, L.~Eslami, S. M. A.and Van~Gool, C.~K.~I. Williams, J.~Winn, and
  A.~Zisserman, ``The pascal visual object classes challenge - a
  retrospective,'' \emph{Int. J. Comput. Vision}, 2014.

\bibitem{babenko2011tpami}
B.~Babenko, M.~H. Yang, and S.~Belongie, ``Robust object tracking with online
  multiple instance learning.'' \emph{IEEE Trans. Pattern Anal. Mach. Intell.},
  vol.~33, no.~8, pp. 1619--1632, 2011.

\bibitem{Wu_TPAMI2010}
H.~Wu, A.~C. Sankaranarayanan, and R.~Chellappa, ``Online empirical evaluation
  of tracking algorithms.'' \emph{IEEE Trans. Pattern Anal. Mach. Intell.},
  vol.~32, no.~8, pp. 1443--1458, 2010.

\bibitem{SanMiguel2012}
J.~SanMiguel, A.~Cavallaro, and J.~Mart\'{i}nez, ``Adaptive on-line performance
  evaluation of video trackers,'' \emph{IEEE Trans. Image Proc.}, vol.~21,
  no.~5, pp. 2812--2823, 2012.

\bibitem{Chu2010}
D.~M. Chu and A.~W.~M. Smeulders, ``Thirteen hard cases in visual tracking,''
  in \emph{IEEE International Conference on Advanced Video and Signal Based
  Surveillance}, 2010.

\bibitem{Jaynes2002}
C.~Jaynes, S.~Webb, R.~Steele, and Q.~Xiong, ``An open development environment
  for evaluation of video surveillance systems,'' in \emph{PETS}, 2002.

\bibitem{Collins2005}
R.~Collins, X.~Zhou, and S.~K. Teh, ``An open source tracking testbed and
  evaluation web site,'' in \emph{Perf. Eval. Track. and Surveillance}, 2005.

\bibitem{Doermann2000}
D.~Doermann and D.~Mihalcik, ``Tools and techniques for video performance
  evaluation,'' in \emph{Proc. Int. Conf. Pattern Recognition}, 2000, pp.
  167--170.

\bibitem{Everingham10}
M.~Everingham, L.~Van~Gool, C.~K.~I. Williams, J.~Winn, and A.~Zisserman, ``The
  pascal visual object classes (voc) challenge,'' \emph{Int. J. Comput.
  Vision}, vol.~88, no.~2, pp. 303--338, Jun. 2010.

\bibitem{Yaakov01}
Y.~Bar-Shalom, X.~R. Li, and T.~Kirubarajan, \emph{Estimation with Applications
  to Tracking and Navigation}.\hskip 1em plus 0.5em minus 0.4em\relax John
  Wiley {\&} Sons, Inc., 2001, ch.~11, pp. 438--440.

\bibitem{AndersonDarling1952}
T.~W. Anderson and D.~A. Darling, ``Asymptotic theory of certain `goodness of
  fit' criteria based on stochastic processes,'' \emph{The Annals of
  Mathematical Statistics}, vol.~23, no.~2, pp. 193--212, 1952.

\bibitem{Demsar2006}
J.~Dem{\v s}ar, ``Statistical comparisons of classifiers over multiple
  datasets,'' \emph{Journal of Machine Learning Research}, vol.~7, pp. 1--30,
  2006.

\bibitem{Navidi2011}
W.~Navidi, \emph{Statistics for Engineers and Scientists}.\hskip 1em plus 0.5em
  minus 0.4em\relax McGraw-Hill, 2011.

\bibitem{Demsar2008}
J.~Dem{\v s}ar, ``On the appropriateness of statistical tests in machine
  learning,'' in \emph{Workshop on Evaluation Methods for Machine Learning
  ICML}, 2008.

\bibitem{Nebehay2014WACV}
G.~Nebehay and R.~Pflugfelder, ``Consensus-based matching and tracking of
  keypoints for object tracking,'' in \emph{Winter Conference on Applications
  of Computer Vision}.\hskip 1em plus 0.5em minus 0.4em\relax IEEE, Mar. 2014.

\bibitem{Maresca2013}
M.~E. Maresca and A.~Petrosino, ``Matrioska: A multi-level approach to fast
  tracking by learning,'' in \emph{Proc. Int. Conf. Image Analysis and
  Processing}, 2013, pp. 419--428.

\bibitem{vojir2011cvww}
T.~Vojir and J.~Matas, ``Robustifying the flock of trackers,'' in \emph{Comp.
  Vis. Winter Workshop}.\hskip 1em plus 0.5em minus 0.4em\relax IEEE, 2011, pp.
  91--97.

\bibitem{Lebeda2013vot}
K.~Lebeda, R.~Bowden, and J.~Matas, ``Long-term tracking through failure
  cases,'' in \emph{Vis. Obj. Track. Challenge VOT2013, In conjunction with
  ICCV2013}, 2013.

\bibitem{2013_DUFFNER}
S.~Duffner and C.~Garcia, ``Pixeltrack: a fast adaptive algorithm for tracking
  non-rigid objects,'' in \emph{Proceedings of the International Conference on
  Computer Vision {(ICCV)}}, 2013, pp. 2480--2487.

\bibitem{Cehovin2013}
L.~\v{C}ehovin, M.~Kristan, and A.~Leonardis, ``{Robust Visual Tracking using
  an Adaptive Coupled-layer Visual Model},'' \emph{IEEE Trans. Pattern Anal.
  Mach. Intell.}, vol.~35, no.~4, pp. 941--953, Apr. 2013.

\bibitem{2014_NAM}
H.~Nam, S.~Hong, and B.~Han, ``Online graph-based tracking,'' in \emph{To
  appear in 2014 European Conference on Computer Vision {ECCV}}, 2014.

\bibitem{2014_CAI}
Z.~Cai, L.~Wen, Z.~Lei, N.~Vasconcelos, and S.~Li, ``Robust deformable and
  occluded object tracking with dynamic graph,'' \emph{{IEEE} Transactions on
  Image Processing}, vol.~23, no.~12, pp. 5497--5509, 2014.

\bibitem{2014_MARESCA_a}
M.~Maresca and A.~Petrosino, ``Clustering local motion estimates for robust and
  efficient object tracking,'' in \emph{Proceedings of the Workshop on Visual
  Object Tracking Challenge, European Conference on Computer Vision {(ECCV)}},
  2014.

\bibitem{Adam2006}
A.~Adam, E.~Rivlin, and I.~Shimshoni, ``\BIBforeignlanguage{English}{{Robust
  Fragments-based Tracking using the Integral Histogram}},'' in
  \emph{\BIBforeignlanguage{English}{CVPR}}, vol.~1.\hskip 1em plus 0.5em minus
  0.4em\relax IEEE Computer Society, Jun. 2006, pp. 798--805.

\bibitem{Felsberg2013vot}
M.~Felsberg, ``Enhanced distribution field tracking using channel
  representations,'' in \emph{Vis. Obj. Track. Challenge VOT2013, In
  conjunction with ICCV2013}, 2013.

\bibitem{2014_FELSBERG}
K.~\"{O}fj\"{a}ll and M.~Felsberg, ``Weighted update and comparison for
  channel{-}based distribution field tracking,'' in \emph{Proceedings of the
  Workshop on Visual Object Tracking Challenge, European Conference on Computer
  Vision {(ECCV)}}, 2014.

\bibitem{sevilla2012cvpr}
L.~Sevilla-Lara and E.~G. Learned-Miller, ``Distribution fields for tracking,''
  in \emph{Comp. Vis. Patt. Recognition}.\hskip 1em plus 0.5em minus
  0.4em\relax IEEE, 2012, pp. 1910--1917.

\bibitem{2012_YIJEONG}
K.~M. Yi, H.~Jeong, S.~W. Kim, and J.~Y. Choi, ``Visual tracking with dual
  modeling,'' in \emph{Proceedings of the 27th Conference on Image and Vision
  Computing New Zealand, {IVCNZ} 12}, 2012, pp. 25--30.

\bibitem{2014_VOJIR}
T.~Vojir, J.~Noskova, and J.~Matas, ``Robust scale-adaptive mean-shift for
  tracking,'' \emph{Pattern Recognition Letters}, 2014.

\bibitem{2014_DANELLJAN}
M.~Danelljan, F.~S. Khan, M.~Felsberg, and J.~{V}an~de {W}eijer, ``Adaptive
  color attributes for real-time visual tracking,'' in \emph{2014 Conference on
  Computer Vision and Pattern Recognition {CVPR}}, 2014.

\bibitem{2012_HENRIQUES}
F.~Henriques, R.~Caseiro, P.~Martins, and J.~Batista, ``Exploiting the
  circulant structure of tracking-by-detection with kernels,'' in \emph{2012
  European Conference on Computer Vision {ECCV}}, 2012.

\bibitem{2001_BRIECHLE}
K.~Briechle and U.~D. Hanebeck, ``Template matching using fast normalized cross
  correlation,'' in \emph{Aerospace/Defense Sensing, Simulation, and Controls,
  International Society for Optics and Photonics}, 2001, pp. 95--102.

\bibitem{comaniciuTPAMI2003}
D.~Comaniciu, V.~Ramesh, and P.~Meer, ``Kernel-based object tracking,''
  \emph{IEEE Trans. Pattern Anal. Mach. Intell.}, vol.~25, no.~5, pp. 564--577,
  2003.

\bibitem{Nummiaro03}
K.~Nummiaro, E.~Koller-Meier, and L.~Van~Gool, ``Color features for tracking
  non-rigid objects,'' \emph{Chinese J. Automation}, vol.~29, no.~3, pp.
  345--355, May 2003.

\bibitem{Perez02}
P.~P{\'{e}}rez, C.~Hue, J.~Vermaak, and M.~Gangnet, ``Color-based probabilistic
  tracking,'' in \emph{Proc. European Conf. Computer Vision}, vol.~1, 2002, pp.
  661--675.

\bibitem{zhang2012eccv}
K.~Zhang, L.~Zhang, and M.~H. Yang, ``Real-time compressive tracking,'' in
  \emph{Proc. European Conf. Computer Vision}, ser. Lecture Notes in Computer
  Science.\hskip 1em plus 0.5em minus 0.4em\relax Springer, 2012, pp. 864--877.

\bibitem{2002_COMANICIU}
D.~Comaniciu and P.~Meer, ``Mean shift: A robust approach toward feature space
  analysis,'' \emph{IEEE Trans. Pattern Anal. Mach. Intell.}, vol.~24, no.~5,
  pp. 603--619, 2002.

\bibitem{2014_DUFFNER}
S.~Duffner and C.~Garcia, ``Exploiting contextual motion cues for visual object
  tracking,'' in \emph{Proceedings of the Workshop on Visual Object Tracking
  Challenge, European Conference on Computer Vision {(ECCV)}}, 2014, pp. 1--12.

\bibitem{hare2011iccv}
S.~Hare, A.~Saffari, and P.~H.~S. Torr, ``Struck: Structured output tracking
  with kernels,'' in \emph{Int. Conf. Computer Vision}, D.~N. Metaxas, L.~Quan,
  A.~Sanfeliu, and L.~J.~V. Gool, Eds.\hskip 1em plus 0.5em minus 0.4em\relax
  IEEE, 2011, pp. 263--270.

\bibitem{Henriques2014}
J.~F. Henriques, R.~Caseiro, P.~Martins, and J.~Batista, ``High-speed tracking
  with kernelized correlation filters,'' \emph{IEEE Trans. Pattern Anal. Mach.
  Intell.}, vol.~1, no.~3, pp. 125--141, 2014.

\bibitem{2014_DANELLJAN_BMVC}
M.~Danelljan, G.~H\"{a}ger, F.~S. Khan, and M.~Felsberg, ``Accurate scale
  estimation for robust visual tracking,'' in \emph{Proceedings of the British
  Machine Vision Conference {BMVC}}, 2014.

\bibitem{2014_LIZHU}
L.~Yang and Z.~Jianke, ``A scale adaptive kernel correlation filter tracker
  with feature integration,'' in \emph{Proceedings of the Workshop on Visual
  Object Tracking Challenge, European Conference on Computer Vision {(ECCV)}},
  2014.

\bibitem{Kristan2005a}
M.~Kristan, J.~Pers, M.~Perse, and S.~Kovacic, ``Bayes spectral entropy-based
  measure of camera focus,'' in \emph{Computer Vision Winter Workshop},
  February 2005, pp. 155--164.

\bibitem{Frey2007}
B.~J. Frey and D.~Dueck, ``Clustering by passing messages between data
  points,'' \emph{Science}, vol. 315, pp. 972--976, 2007.

\bibitem{KristanVOT2013}
M.~Kristan, R.~Pflugfelder, A.~Leonardis, J.~Matas, F.~Porikli, L.~Cehovin,
  G.~Nebehay, G.~Fernandez, T.~Vojir, and et~al., ``The visual object tracking
  vot2013 challenge results,'' in \emph{ICCV2013 Workshops, Workshop on visual
  object tracking challenge}, 2013, pp. 98 --111.

\bibitem{Bolme2010}
D.~S. Bolme, J.~R. Beveridge, B.~A. Draper, and Y.~M. Lui, ``Visual object
  tracking using adaptive correlation filters,'' in \emph{Comp. Vis. Patt.
  Recognition}, 2010.

\bibitem{Dalal05}
N.~Dalal and B.~Triggs, ``Histograms of oriented gradients for human
  detection,'' in \emph{Comp. Vis. Patt. Recognition}, vol.~1, June 2005, pp.
  886--893.

\bibitem{Torralba_CVPR2011}
A.~Torralba and A.~A. Efros, ``Unbiased look at dataset bias,'' in \emph{Comp.
  Vis. Patt. Recognition}.\hskip 1em plus 0.5em minus 0.4em\relax IEEE, 2011,
  pp. 1521--1528.

\bibitem{Li2015tpami}
A.~Li, M.~Li, Y.~Wu, M.~Yang, and S.~Yan, ``Nus-pro: A new visual tracking
  challenge,'' \emph{IEEE Trans. Pattern Anal. Mach. Intell.}, 2015.

\end{thebibliography}

\vspace{-1cm}
\begin{IEEEbiography}[{\includegraphics[width=1in,height=1.25in,clip,keepaspectratio]{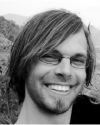}}]{Matej Kristan}
received a Ph.D from the Faculty of Electrical Engineering, University of Ljubljana in 2008. He is an Assistant Professor at the ViCoS Laboratory at the Faculty of Computer and Information Science and at the Faculty of Electrical Engineering, University of Ljubljana. His research interests include probabilistic methods for computer vision with focus on visual tracking, dynamic models, online learning, object detection and vision for mobile robotics.
\end{IEEEbiography} \vspace{-1cm}

\begin{IEEEbiography}[{\includegraphics[width=1in,height=1.25in,clip,keepaspectratio]{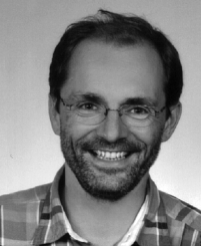}}]{Jiri Matas}
is a Professor at the Center for Machine Perception, Faculty of Electrical Engineering at the Czech Technical University in Prague, Czech Republic. He is author or co-author of more than~$250$ papers in the area of Computer Vision and Machine Learning. His research interests include object recognition, image retrieval, tracking, sequential pattern recognition, invariant feature detection and Hough Transform and RANSAC-type optimization.
\end{IEEEbiography} \vspace{-1cm}

\begin{IEEEbiography}[{\includegraphics[width=1in,height=1.25in,clip,keepaspectratio]{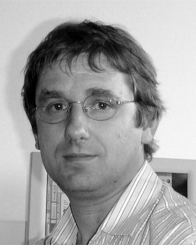}}]{Ale{\v s} Leonardis}
is a Professor at the School of Computer Science, University of Birmingham and co-Director of the Centre for Computational Neuroscience
and Cognitive Robotics. He is also a Professor at the FCIS, University of Ljubljana and adjunct professor at the FCS, TU-Graz. His research interests include robust and adaptive methods for computer vision, object and scene recognition and categorization, statistical visual learning, 3D object modeling, and biologically motivated vision.
\end{IEEEbiography} \vspace{-1cm}

\begin{IEEEbiography}[{\includegraphics[width=1in,height=1.25in,clip,keepaspectratio]{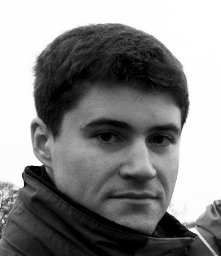}}]{Tom{\' a}{\v s} Voj{\' i}{\v r} }
received a Master degree at the Faculty of Electrical Engineering at the Czech Technical University in Prague in 2010 and is currently pursuing his PhD at the same faculty at the Center for Machine Perception. His research interests are computer vision with focus in real-time visual object tracking. 
\end{IEEEbiography} \vspace{-1cm}

\begin{IEEEbiography}[{\includegraphics[width=1in,height=1.25in,clip,keepaspectratio]{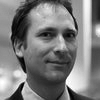}}]{Roman Pflugfelder}
received his Ph.D. degree at the Institute of Computer Graphics at Graz University of Technology in 2008. Currently, Dr Pflugfelder is scientist and project leader at the Austrian Institute of Technology. His research interests lie in object tracking and camera auto-calibration for video surveillance.
\end{IEEEbiography} \vspace{-1cm}

\begin{IEEEbiography}[{\includegraphics[width=1in,height=1.25in,clip,keepaspectratio]{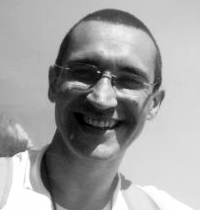}}]{Gustavo Fernandez}
is scientist at the Video and Security Technology Group, Austrian Institute of Technology. He obtained his master degree from the University of Buenos Aires (Argentina) in 2000 and his PhD from Graz University of Technology (Austria) in 2004. Since 2005 he is at the Austrian Institute of Technology doing research in the area of Computer Vision. His research interests are both theory and applications of object detection, object tracking and cognitive systems.
\end{IEEEbiography} \vspace{-1cm}
 
\begin{IEEEbiography}[{\includegraphics[width=1in,height=1.25in,clip,keepaspectratio]{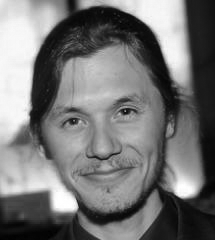}}]{Georg Nebehay}
received his Master Degree in Computer Science in 2011 from Vienna University of Technoogy, Austria. He is currently PhD student at the Austrian Institute of Technology. His research interests are tracking and video surveillance.
\end{IEEEbiography} \vspace{-1cm}

\begin{IEEEbiography}[{\includegraphics[width=1in,height=1.25in,clip,keepaspectratio]{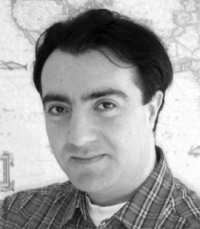}}]{Fatih Porikli}
is a Professor at the Australian National University in Canberra. Previously, Professor Porikli was a Distinguished Research Scientist at Mitsubishi Electric Research Labs (MERL). He received his PhD degree, with specialization in video object segmentation, from NYU Poly, USA. He has authored more than 100 publications and invented more than 60 patents.
His work covers areas including computer vision, machine learning, video surveillance, multimedia processing among others.
\end{IEEEbiography} \vspace{-1cm}

\begin{IEEEbiography}[{\includegraphics[width=1in,height=1.25in,clip,keepaspectratio]{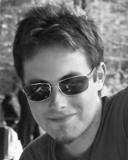}}]{Luka {\v C}ehovin} received his Ph.D from the Faculty of Computer and Information Science, University of Ljubljana, Slovenia in 2015.
Currently he is working at the Visual Cognitive Systems Laboratory,
Faculty of Computer and Information Science, University of Ljubljana,
Slovenia as a teaching assistant and a researcher.
His research interests include computer vision, HCI, distributed
intelligence and web-mobile technologies.
\end{IEEEbiography}\vspace{-1cm}

\end{document}